\documentclass[onefignum,onetabnum]{siamart190516}

\usepackage{lipsum}

\usepackage{amsfonts}
\usepackage{textcomp}

\ifpdf
  \DeclareGraphicsExtensions{.eps,.pdf,.png,.jpg}
\else
  \DeclareGraphicsExtensions{.eps}
\fi

\usepackage{enumitem}
\setlist[enumerate]{leftmargin=.5in}
\setlist[itemize]{leftmargin=.5in}

\newsiamremark{remark}{Remark}
\newsiamremark{hypothesis}{Hypothesis}
\crefname{hypothesis}{Hypothesis}{Hypotheses}
\newsiamthm{claim}{Claim}
\newsiamthm{example}{Example}

\headers{An Example Article}{D. Doe, P. T. Frank, and J. E. Smith}

\title{Accelerating Stochastic Optimization of Separable Deep Neural Networks via Iterative Sampling Methods\thanks{Submitted to the editors DATE.
\funding{This work was funded by the Fog Research Institute under contract no.~FRI-454.}}}

\author{Dianne Doe\thanks{Imagination Corp., Chicago, IL 
  (\email{ddoe@imag.com}, \url{http://www.imag.com/\string~ddoe/}).}
\and Paul T. Frank\thanks{Department of Applied Mathematics, Fictional University, Boise, ID 
  (\email{ptfrank@fictional.edu}, \email{jesmith@fictional.edu}).}
\and Jane E. Smith\footnotemark[3]}

\usepackage{amsopn}

\usepackage{tikz}
\usepackage{multirow}

\DeclareMathOperator*{\argmin}{arg\: min}

\makeatletter
\newcommand*{\addFileDependency}[1]{% argument=file name and extension
  \typeout{(#1)}% latexmk will find this if $recorder=0 (however, in that case, it will ignore #1 if it is a .aux or .pdf file etc and it exists! if it doesn't exist, it will appear in the list of dependents regardless)
  \@addtofilelist{#1}% if you want it to appear in \listfiles, not really necessary and latexmk doesn't use this
  \IfFileExists{#1}{}{\typeout{No file #1.}}% latexmk will find this message if #1 doesn't exist (yet)
}
\makeatother

\usepackage{array}
\usepackage{arydshln}

\def\mydefb#1{\expandafter\def\csname bf#1\endcsname{\mathbf{#1}}}
\def\mydefallb#1{\ifx#1\mydefallb\else\mydefb#1\expandafter\mydefallb\fi}
\mydefallb aAbBcCdDeEfFgGhHiIjJkKlLmMnNoOpPqQrRsStTuUvVwWxXyYzZ\mydefallb

\def\mydefgreek#1{\expandafter\def\csname bf#1\endcsname{\text{\boldmath$\mathbf{\csname #1\endcsname}$}}}
\def\mydefallgreek#1{\ifx\mydefallgreek#1\else\mydefgreek{#1}%
   \lowercase{\mydefgreek{#1}}\expandafter\mydefallgreek\fi}
\mydefallgreek {alpha}{Alpha}{beta}{Beta}{gamma}{Gamma}{delta}{Delta}{epsilon}{Epsilon}{zeta}{Zeta}{eta}{Eta}{theta}{Theta}{iota}{Iota}{kappa}{Kappa}{lambda}{Lambda}{mu}{Mu}{nu}{Nu}{omicron}{Omicron}{pi}{Pi}{rho}{Rho}{sigma}{Sigma}{tau}{Tau}{upsilon}{Upsilon}{phi}{Phi}{xi}{Xi}{chi}{Chi}{psi}{Psi}{omega}{Omega}\mydefallgreek

\def\mydefb#1{\expandafter\def\csname bb#1\endcsname{\mathbb{#1}}}
\def\mydefallb#1{\ifx#1\mydefallb\else\mydefb#1\expandafter\mydefallb\fi}
\mydefallb aAbBcCdDeEfFgGhHiIjJkKlLmMnNoOpPqQrRsStTuUvVwWxXyYzZ\mydefallb

\def\mydefb#1{\expandafter\def\csname #1bb\endcsname{\mathbb{#1}}}
\def\mydefallb#1{\ifx#1\mydefallb\else\mydefb#1\expandafter\mydefallb\fi}
\mydefallb aAbBcCdDeEfFgGhHiIjJkKlLmMnNoOpPqQrRsStTuUvVwWxXyYzZ\mydefallb

\def\mydefb#1{\expandafter\def\csname #1cal\endcsname{\mathcal{#1}}}
\def\mydefallb#1{\ifx#1\mydefallb\else\mydefb#1\expandafter\mydefallb\fi}
\mydefallb aAbBcCdDeEfFgGhHiIjJkKlLmMnNoOpPqQrRsStTuUvVwWxXyYzZ\mydefallb

\def\mydefb#1{\expandafter\def\csname cal#1\endcsname{\mathcal{#1}}}
\def\mydefallb#1{\ifx#1\mydefallb\else\mydefb#1\expandafter\mydefallb\fi}
\mydefallb aAbBcCdDeEfFgGhHiIjJkKlLmMnNoOpPqQrRsStTuUvVwWxXyYzZ\mydefallb

\newcommand{\nFeatIn}{n_{\rm in}} %dimension of y
\newcommand{\nFeatOut}{n_{\rm out}} %dimension of F(y,theta)
\newcommand{\nTargets}{n_{\rm target}} %dimension of c
 
\newcommand{\nWeights}{n_{\theta}} % dimension of theta
\newcommand{\nTheta}{\nWeights} % dimension of theta
\newcommand{\batchSize}{|\Tcal_k|} % batch size
\newcommand{\memDepth}{r}

\renewcommand{\t} {^{\top}}                             
\newcommand{\bfzero}{{\bf0}}
\newcommand{\norm} [2][]{\left\|#2\right\|_{#1}}        
                                
\newcommand{\trace} [1]  {{\rm tr\!}\left( #1 \right)}
\newcommand{\mdot} {\,\cdot\,}

\newcommand{\thf}  {\tfrac{1}{2}}                            % 1/2

\usepackage{lipsum}
\usepackage{amsfonts}
\usepackage{epstopdf}
\RequirePackage{algpseudocode}
\ifpdf
  \DeclareGraphicsExtensions{.eps,.pdf,.png,.jpg}
\else
  \DeclareGraphicsExtensions{.eps}
\fi

\usepackage{enumitem}
\setlist[enumerate]{leftmargin=.5in}
\setlist[itemize]{leftmargin=.5in}

\headers{{{ slimTrain}} - A Stochastic Approximation Method for Separable DNNs}{Newman, Chung, Chung, and Ruthotto}

\title{
slimTrain - A Stochastic Approximation Method for Training Separable Deep Neural Networks\thanks{Submitted to the editors \today.
\funding{This work was partially supported by the National Science Foundation (NSF) under grant DMS-1654175 (J. Chung), DMS-1723005 (M. Chung and J. Chung), and DMS-1751636 (Ruthotto), Air Force Office of Scientific Research   Grant   20RT0237 (Newman and Ruthotto), the US Department of Energy's  Office  of  Advanced  Scientific  Computing  Research  Field  Proposal 20-023231 (Newman and Ruthotto)}}}

\author{Elizabeth Newman\thanks{Department of Mathematics, Emory University, Atlanta, GA 
  (\email{elizabeth.newman@emory.edu}, \url{http://math.emory.edu/\string~enewma5}).}
\and Julianne Chung\thanks{Department of Mathematics, Academy of Data Science, Virginia Tech, Blacksburg, VA
  (\email{jmchung@vt.edu}, \url{https://intranet.math.vt.edu/people/jmchung}).}
  \and Matthias Chung\thanks{Department of Mathematics, Academy of Data Science, Virginia Tech, Blacksburg, VA
  (\email{mcchung@vt.edu}, \url{https://intranet.math.vt.edu/people/mcchung}).}
\and Lars Ruthotto\thanks{Department of Mathematics, Emory University, Atlanta, GA 
  (\email{lruthotto@emory.edu},\url{http://www.mathcs.emory.edu/\string~lruthot})}}

\ifpdf
\hypersetup{
  pdftitle={},
  pdfauthor={}
}
\fi

\begin{document}

\maketitle

% REQUIRED
\begin{abstract}
Deep neural networks (DNNs) have shown their success as high-dimensional function approximators in many applications; however, training DNNs can be challenging in general. DNN training is commonly phrased as a stochastic optimization problem whose challenges include non-convexity, non-smoothness, insufficient regularization, and complicated data distributions. Hence, the performance of DNNs on a given task depends crucially on tuning hyperparameters, especially learning rates and regularization parameters. In the absence of theoretical guidelines or prior experience on similar tasks, this requires solving a series of repeated training problems which can be time-consuming and demanding on computational resources. This can limit the applicability of DNNs to problems with non-standard, complex, and scarce datasets, e.g., those arising in many scientific applications. To remedy the challenges of DNN training, we propose {\tt slimTrain}, a stochastic optimization method for training DNNs with reduced sensitivity to the choice hyperparameters and fast initial convergence. The central idea of {\tt slimTrain} is to exploit the separability inherent in many DNN architectures; that is, we separate the DNN into a nonlinear feature extractor followed by a linear model. This separability allows us to leverage recent advances made for solving large-scale, linear, ill-posed inverse problems. Crucially, for the linear weights, {\tt slimTrain} does not require a learning rate and automatically adapts the regularization parameter.  In our numerical experiments using function approximation tasks arising in surrogate modeling and dimensionality reduction, {\tt slimTrain} outperforms existing DNN training methods with the recommended hyperparameter settings and reduces the sensitivity of DNN training to the remaining hyperparameters. Since our method operates on mini-batches, its computational overhead per iteration is modest and savings can be realized by reducing the number of iterations (due to quicker initial convergence) or the number of training problems that need to be solved to identify effective hyperparameters.
\end{abstract}

% REQUIRED

% REQUIRED

% ====================================================================== %
% Section 1 - Introduction
% ====================================================================== %

\section{Introduction}
Deep neural networks (DNNs) provide a powerful framework for approximating complex mappings, possessing universal approximation properties~\cite{Cybenko1989} and flexible architectures composed of simple functions parameterized by weights. Numerous studies have shown that excellent performance can be obtained using state-of-the-art DNNs in numerous applications including mage processing, speech recognition, surrogate modeling, and dimensionality reduction \cite{Goodfellow:2016wc,RuthottoHaber2019,Tripathy:2018kk}. However, getting such results \textit{in practice} may be a computationally expensive and cumbersome task. The process of training DNNs, or finding the optimal weights, is rife with challenges, e.g., the optimization problem is non-convex, expressive networks require a very large number of weights, and, perhaps most critically, appropriate regularization is needed to ensure the trained network generalizes well to unseen data.  Due to these challenges, it can be difficult to train a network efficiently and to sufficient accuracy, especially for large, high-dimensional datasets and complex mappings and in the absence of experience on similar learning task. This is a particular challenge in scientific applications that often involve unique training data sets, which limits the use of standard architectures and established hyperparameters.

While the literature on effective solvers for training DNNs is vast (see, e.g., the recent survey~\cite{bottou2016optimization}), the most popular approaches are stochastic approximation (SA) methods. SA methods are computationally appealing since only a small, randomly-chosen sample (i.e., mini-batch) from the training data is needed at each iteration to update the DNN parameters.  Also, SA methods tend to exhibit good generalization properties. The most extensively studied and utilized SA method is the stochastic gradient descent (SGD) method \cite{RobbinsMonro1951} and its many popular variants such as AdaGrad \cite{duchi2011adaptive} and ADAM \cite{kingma2014adam}. Despite the popularity of SGD variants, major disadvantages include slow convergence and, most notoriously, the need to select a suitable learning rate (step size).  Stochastic Newton and stochastic quasi-Newton methods have been proposed to accelerate convergence of SA methods \cite{bottou2004large,gower2017randomized,Byrd2016,wang2017stochastic, chung2017stochastic}, but including curvature information in SA methods is not trivial.  Contrary to deterministic methods, which are known to benefit from the use of second-order information (consider, e.g., the natural step size of one and local quadratic convergence of Newton's method), noisy curvature estimates in stochastic methods may have harmful effects on the robustness of the iterations \cite{chung2017stochastic}.  Furthermore, SA methods cannot achieve a convergence rate that is faster than sublinear \cite{agarwal2012information}, and additional care must be taken to handle nonlinear, nonconvex problems arising in DNN training. The performance and convergence properties of SA methods depend heavily on the properties of the objective function and on the choice of the learning rate.

In this paper, we seek to simplify the training of DNNs by exploiting the \emph{separability} inherent in most common DNN architectures. We assume that the network, $G$, is parameterized by two blocks of weights, $\bfW$ and $\bftheta$, and is of the form
	\begin{equation}\label{eq:separableDNN}
    	G(\cdot, \bfW, \bftheta) = \bfW F(\cdot, \bftheta),
	\end{equation}
where $F$, also referred to as a feature extractor, is a parameterized, nonlinear function. The important observation here is that the DNN is nonlinear in $\bftheta$ and, crucially, is \emph{linear in $\bfW$}.  Any DNN whose last layer does not contain a nonlinear activation function can be written in this form, so our definition includes many state-of-the-art DNNs; see, e.g.,~\cite{he2016deep, Raissi:2019hv, Lecun:1990mnist, Krizhevsky:2012alexnet, ronneberger2015unet} and following works like \cite{Sjoberg1997,Tripathy:2018kk,newman2020train}. In a supervised learning framework, the goal is to find a set of network weights, $(\bfW, \bftheta)$, such that $\bfW F(\bfy,\bftheta) \approx \bfc$ for all input-target pairs $(\bfy,\bfc)$ in a data space. Training the network means learning the network weights by minimizing an expected loss or discrepancy of the DNN approximation over all input-target pairs $(\bfy, \bfc)$ in a training set, while generalizing well to unobserved input-target pairs. 

\paragraph{\textit{\textbf{Main contributions}}}
In this paper, we describe {\tt slimTrain}, a \textit{sampled limited-memory training} method that  exploits the separability of the DNN architecture to leverage recently-developed sampled Tikhonov methods for automatic regularization parameter tuning \cite{newman2020train, chung2020sampled}.For the linear weights in a regression framework, we obtain a stochastic linear least-squares problem, and we use recent work on sampled limited-memory methods to approximate the global curvature of the underlying least-squares problem.  Such methods can be viewed as row-action or SA methods and can speed up the initial convergence and to improve the accuracy of iterates \cite{chung2020sampled}.  As discussed above, applying a second-order SA method to the entire problem is not trivial and obtaining curvature information for the nonlinear weights is computationally expensive, particularly for deep networks.  As our approach only incorporates curvature in the final layer of the network, where we have a linear structure, its computational overhead is minimal.  In doing so, we can not only improve initial convergence of DNN training, but also can select the regularization parameter automatically by exploiting connections between the learning rate of the linear weights and the regularization parameter for Tikhonov regularization \cite{chung2019iterative}. Thus, {\tt slimTrain} is an efficient, practical method for training separable DNNs that is memory-efficient (i.e., working only on mini-batches), exhibits faster initial convergence compared to standard SA approaches (e.g., ADAM), produces networks that generalize well, and incorporates automatic hyperparameter selection.

The paper is organized as follows. In \cref{sec:separable}, we describe separable DNN architectures and review various approaches to train such networks, with special emphasis on variable projection.  Notably, ae provide new theoretical analysis to support a VarPro stochastic approximation method. In \cref{sec:sa}, we introduce our new {\tt slimTrain} approach that incorporates sampled limited-memory Tikhonov ({\tt slimTik}) methods within the nonlinear learning problem.  Here, we describe cross-validation-based techniques to automatically and adaptively select the regularization parameter.  Numerical results are provided in \cref{sec:results}, and conclusions follow in \cref{sec:conclusions}.

% ====================================================================== %
% Section 2 - Separability
% ====================================================================== %

\section{Exploiting separability with variable projection}
\label{sec:separable}
Given the space of input features $\Ycal \subseteq \Rbb^{\nFeatIn}$ and the space of target features $\Ccal \subseteq \Rbb^{\nTargets}$, let $\Dcal \subseteq \Ycal \times \Ccal$ be the data space containing input-target pairs $(\bfy, \bfc) \in \Dcal$. We focus on separable DNN architectures that consist of two separate phases: a nonlinear feature extractor $F: \Ycal \times \Rbb^{\nWeights} \to \Rbb^{\nFeatOut}$ parametrized by $\bftheta\in \Rbb^{\nWeights}$ followed by a linear model $\bfW \in \Rbb^{\nTargets\times \nFeatOut}$. In general, the goal is to learn the network weights, $(\bfW,\bftheta)$, by solving the stochastic optimization problem
    \begin{align}\label{eq:objFull}
        \min_{\bfW, \bftheta}  \  \Ebb  \ L(\bfW F(\bfy, \bftheta), \bfc) + R(\bftheta) + S(\bfW),
    \end{align}
where $L:\Rbb^{\nTargets} \times \Ccal \to \Rbb$ is a loss function, and $R:\Rbb^{\nWeights} \to \Rbb$ and $S:\Rbb^{\nTargets \times \nFeatOut} \to \Rbb$ are regularizers.  Here, $\Ebb$ denotes the expected value over a distribution of input-target pairs in $\Dcal$. 

Choosing an appropriate loss function $L$ is task-dependent.  For example, a least-squares loss function promotes data-fitting and is well-suited for function approximation tasks whereas a cross-entropy loss function is preferred for classification tasks where the network outputs are interpreted as a discrete probability distribution~\cite{hui2020evaluation}. In this work, we focus on exploiting separability to improve DNN training for function approximation or data fitting tasks such as PDE surrogate modeling~\cite{Tripathy:2018kk,Zhu:2018ik} and dimensionality reduction such as autoencoders~\cite{Goodfellow:2016wc}.  Hence, we restrict our focus to a stochastic least-squares loss function with Tikhonov regularization
    \begin{align}
    \label{eq:objFctnLSFull}
        \min_{\bfW,\bftheta} \ \Phi(\bfW,\bftheta) &\equiv 
             \Ebb \  \thf \left\|\bfW F(\bfy, \bftheta) - \bfc\right\|_2^2 +  \tfrac{\alpha}{2}\|\bfL\bftheta\|_2^2+ \tfrac{\lambda}{2}\|\bfW\|_{\rm F}^2,
    \end{align}
  where $\Phi: \Rbb^{\nTargets \times \nFeatOut} \times \Rbb^{\nWeights} \to \Rbb$ is the objective function, $\bfL$ is a user-defined operator, $\|\cdot \|_{\rm F}$ is the Frobenius norm, and $\alpha, \lambda \geq 0$ are the regularization parameters for $\bftheta$ and $\bfW$, respectively.  

% --------------------------------------------------------------------------------------------------------------------------- %
\subsection{SA methods that exploit separability}
A standard, and the current state-of-the-art, approach to solve~\eqref{eq:objFctnLSFull} is stochastic optimization over both sets of weights $(\bfW,\bftheta)$ simultaneously (i.e., joint estimation). While generic and straightforward, this fully-coupled approach can suffer from slow convergence (e.g., due to ill-conditioning) and does not attain potential benefits that can be achieved by treating the separate blocks of weights differently (e.g., exploiting the structure of the arising subproblems). We seek computational methods for training DNNs that exploit separability, i.e., we treat the two parameter sets $\bftheta$ and $\bfW$ differently and exploit linearity in $\bfW$.  Three general approaches to numerically tackle the optimization problem~\cref{eq:objFctnLSFull} while taking advantage of the separability are as follows. 

\paragraph{\textit{\textbf{Alternating directions}}} 
One approach that exploits separability of the variables $\bftheta$ and $\bfW$ is alternating optimization~\cite{bezdek2002some}. For~\cref{eq:objFctnLSFull}, this corresponds to alternating between two stochastic optimization problems. Note for simplicity of presentation we assume that each of following optimization problems has a unique minimizer. Suppose we initialize $\bftheta_0$. Then, at the $k$-th iteration, we embark on
    \begin{equation} 
    \label{eq:optW}
        \bfW_k = \argmin_{\bfW} \ \Phi(\bfW, \bftheta_{k-1})
    \end{equation}
    and
    \begin{equation} 
    \label{eq:opttheta}
        \bftheta_k = \argmin_{\bftheta} \ 
             \ \Phi(\bfW_{k}, \bftheta).
    \end{equation}
    Notice that convergence of this approach can be slow when variables are tightly coupled~\cite{Beck2013:bcgd, wright2015coordinate}. Furthermore, this approach is not practical in our settings, since minimization problem~\cref{eq:optW} and \cref{eq:opttheta} are computationally expensive, particularly the non-convex, high-dimensional, often non-smooth optimization problem for $\bftheta$.  
    
\paragraph{\textit{\textbf{Block coordinate descent}}} 
A practical alternative for alternating directions is block coordinate descent. The general idea of a block coordinate descent approach for \cref{eq:objFctnLSFull} is to approximate the alternating optimization of \cref{eq:optW} and \cref{eq:opttheta} via iterative update schemes (e.g., one iteration of an iterative optimization step) for each set of variables~\cite{wright2015coordinate}.  Note that under certain assumptions, a block coordinate descent method applied to two sets of parameters has been shown to converge~\cite{nesterov2012efficiency, richtarik2014iteration}.  Although a block coordinate descent approach provides a computationally appealing alternative to the fully coupled and alternating directions approaches, this approach, like alternating directions, suffers from slow convergence when the blocks are tightly coupled.

\paragraph{\textit{\textbf{Variable projection (VarPro)}}} 
A compromise between alternating directions and block coordinate descent is to solve~\cref{eq:optW} with respect to $\bfW$ while performing an iterative update method for \cref{eq:opttheta} with respect to $\bftheta$. This can be seen as a stochastic approximation version of a variable projection approach~\cite{GolubPereyra2003}. Formally, we can write the iteration in terms of the  \textit{reduced} stochastic optimization problem
        \begin{align} 
        \label{eqn:objFctnReducedphi}
        \min_{\bftheta} \ \Phi^{\rm red}(\bftheta) &\equiv \Phi(\widehat{\bfW}(\bftheta),\bftheta)
    \end{align}
    where 
    \begin{equation} 
    \label{eq:objFctnLinphi}
        \widehat\bfW(\bftheta) = \argmin_{\bfW}  \Ebb \ \thf \left\|\bfW F(\bfy,\bftheta) - \bfc\right\|_2^2 +  \tfrac{\lambda}{2}\|\bfW\|_{\rm F}^2.
    \end{equation}
Notice that~\eqref{eq:objFctnLinphi} is a \textit{stochastic} Tikhonov-regularized \textit{linear} least-squares problem and, under mild assumptions, there exists a closed form solution, i.e.,
 \begin{equation}\label{eq:What}
       \widehat \bfW(\bftheta) = \bbE \bfc F(\bfy,\bftheta)\t \left(\bfSigma_{\bfy}(\bftheta)  +  \bfmu_\bfy(\bftheta)\bfmu_\bfy(\bftheta)\t  + \lambda\bfI\right)^{-1}.
\end{equation}
Here, $\bfmu_\bfy(\bftheta) = \bbE F(\bfy,\bftheta)$ and $\bfSigma_\bfy(\bftheta) = \bbE (F(\bfy,\bftheta) -\bfmu_\bfy)(F(\bfy,\bftheta) -\bfmu_\bfy)\t$. Details of the derivation can be found in \cref{sec:stochlinearTik}. 

% --------------------------------------------------------------------------------------------------------------------------- %
\subsection{Theoretical justification for VarPro in SA methods}
\label{sub:updatetheta}

After solving for $\widehat{\bfW}(\bftheta)$ in~\eqref{eq:objFctnLinphi}, VarPro uses an iterative  scheme, typically an SGD variant, to update $\bftheta$. The key is to ensure that the mini-batch gradients used to update $\bftheta$ are unbiased. To the best of our knowledge, we provide the first theoretical analysis demonstrating that VarPro in an SA setting produces an unbiased estimate of the gradient.  We note that the derivation, presented for stochastic Tikhonov-regularized least-squares problems, can be extended to any objective function which is convex with respect to the linear weights, such as when using a cross-entropy loss function.

In the context of the DNN training problem, let $\Tcal\subseteq \Dcal$ be a finite training set.  At the $k$-th training iteration, we select a mini-batch the training set, $\Tcal_k\subset \Tcal$.   For the $\Tcal_k$ we seek to minimize the function
    \begin{equation}\label{eq:batchObjFctn}
        \Phi_k(\bfW, \bftheta) = \frac{1}{|\Tcal_k|}\sum_{(\bfy,\bfc)\in \Tcal_k}\tfrac{1}{2}\norm[2]{\bfW F(\bfy,\bftheta) - \bfc}^2 + \tfrac{\alpha}{2}\norm[2]{\bfL\bftheta}^2 + \tfrac{\lambda}{2}\norm[\rm F]{\bfW}^2.
    \end{equation}
A VarPro SA method applied to \cref{eqn:objFctnReducedphi} considers the reduced functional at the $k$-th iteration,
    \begin{align}
        \Phi_k^{\rm red}(\bftheta) & = \Phi_k(\widehat\bfW(\bftheta), \bftheta)
    \end{align}
   where $\widehat{\bfW}(\bftheta)$ is obtained from~\eqref{eq:objFctnLinphi}, i.e., the solution to the stochastic Tikhonov-regularized linear least-squares problem \emph{over the entire data space}.  

   To update the nonlinear weights, we select a ``descent'' direction $\bfp_k$ with respect to $\bftheta$ and compute the next iterate,
    \begin{equation}
        \bftheta_k  = \bftheta_{k-1} + \gamma \bfp_k(\bftheta_{k-1};\widehat\bfW(\bftheta_{k-1})).
    \end{equation}
    Here, $\gamma$ denotes an appropriate learning rate and $\bfp_k$ is a direction that is computed based on the current estimate of $\bftheta_{k-1}$ with respect to the current batch $\calT_k$. The selection of $\bfp_k$ depends on the chosen stochastic optimization method and requires knowing information about the derivative of~\eqref{eq:batchObjFctn}.  Explicitly, we compute the derivative of $\Phi^{\rm red}_k$ with respect to $\bftheta$ by 
       \begin{equation}
       \begin{split}
        {\rm D}_\bftheta  \Phi_k^{\rm red}(\bftheta) &= {\rm D}_\bftheta  \Phi_k (\widehat\bfW(\bftheta),\bftheta) \\  
         & = \left[{\rm D}_{\bfW} \Phi_k(\bfW,\bftheta)\right]_{\bfW = \widehat\bfW(\bftheta)}  \cdot  {\rm D}_{\bftheta}\widehat\bfW(\bftheta) + \left[{\rm D}_{\widetilde \bftheta} \Phi_k(\widehat\bfW(\bftheta),\widetilde \bftheta)\right]_{\widetilde\bftheta = \bftheta}. \label{eq:Varpro_grad}
        \end{split}
    \end{equation} 
     Note that, contrary to VarPro derivations in deterministic settings \cite{GolubPereyra2003, chung2010efficient, newman2020train}, the first term in \cref{eq:Varpro_grad} does not vanish. This is because $\widehat\bfW(\bftheta)$ satisfies the optimality conditions for $\Phi$, the objective function for expected value minimization problem \cref{eq:objFctnLinphi} but may not be optimal for $\Phi_k$, the objective function for the current batch.  However, we observe that the term vanishes in expectation over all samples, that is,
    \begin{equation}\label{eq:unbiasedVarPro}
    \begin{split}
        \bbE \,\left( \left[{\rm D}_{\bfW} \Phi_k(\bfW,\bftheta)\right]_{\bfW = \widehat\bfW(\bftheta)}  \cdot  {\rm D}_{\bftheta}\widehat\bfW(\bftheta) \right) 
        & =  \left[{\rm D}_{\bfW} \bbE \,\Phi_k(\bfW,\bftheta)\right]_{\bfW = \widehat\bfW(\bftheta)}  \cdot  {\rm D}_{\bftheta}\widehat\bfW(\bftheta)  \\
        & =
         \left[{\rm D}_{\bfW} \Phi(\bfW,\bftheta)\right]_{\bfW = \widehat\bfW(\bftheta)}  \cdot  {\rm D}_{\bftheta}\widehat\bfW(\bftheta) \\
         &=
        \bfzero.
       \end{split}
    \end{equation}
    Because~\eqref{eq:Varpro_grad} is equal to the gradient of the full objective function $\Phi$ in expectation, we say the update for $\bftheta$ is unbiased.
    Since SA methods can handle unbiased noisy gradients, one could define a VarPro SGD approach using the following unbiased estimator for the gradient,
    \begin{equation}\label{eq:direction}
        \bfp_k(\bftheta; \widehat\bfW(\bftheta)) = -\left[{\rm D}_{\widetilde\bftheta} \Phi_k(\widehat\bfW(\bftheta),\widetilde\bftheta)\right]_{\widetilde\bftheta = \bftheta}\t
    \end{equation} 
    where the derivative is
    \begin{equation}
    \label{eq:p_approx} 
    \begin{split}
        {\rm D}_{\bftheta} \Phi_k(\bfW,\bftheta)
        &={\rm D}_{\bftheta}\left(\frac{1}{|\Tcal_k|}\sum_{(\bfy,\bfc)\in \Tcal_k}\tfrac{1}{2}\norm[2]{\bfW F(\bfy,\bftheta) - \bfc}^2 + \tfrac{\alpha}{2}\norm[2]{\bfL\bftheta}^2 \right)\\
         &=\frac{1}{|\Tcal_k|}\sum_{(\bfy,\bfc)\in \Tcal_k} \left(\bfW F(\bfy,\bftheta)\t - \bfc\right)\bfW{\rm D}_{\bftheta} F(\bfy,\bftheta)
         +\alpha\bftheta\t\bfL\t\bfL. 
    \end{split}
    \end{equation} 
    Note that ${\rm D}_{\bftheta} F(\bfy,\bftheta)$ can be obtained through back propagation which can be parallelized over samples.
    
% --------------------------------------------------------------------------------------------------------------------------- %
\subsection{Challenges of VarPro in stochastic optimization}
\label{sub:challengesVarPro}

The appeal of a VarPro approach is marred by the impracticality of computing $\widehat\bfW(\bftheta)$ in~\eqref{eq:objFctnLinphi}. For each mini-batch update of $\bftheta$, one would need to recompute $\widehat\bfW(\bftheta)$, which requires propagating many samples through the network.  Since a computation is costly, in terms of time and storage, we can only obtain an approximation of $\widehat{\bfW}(\bftheta)$ in practice.

One way to approximate $\widehat\bfW(\bftheta)$ is to replace the vector $\bfmu_\bfy(\bftheta)$ and the matrix $\bbE \bfc F(\bfy,\bftheta)\t$ with sample mean approximations and the covariance matrix $\bfSigma_\bfy(\bftheta)$ with a sample covariance matrix.  The accuracy of the approximation, and hence the expected bias of the gradients for the nonlinear weights, will depend on the size of the sample. However, these quantities still depend on $\bftheta$, and hence for any iterative process where $\bftheta$ is being updated, these values need to be recomputed at each iteration. 

A practical strategy to approximate $\widehat\bfW(\bftheta)$ is to use a sample average approximation (SAA) approach.  In SAA methods, one first approximates the expected loss using a (large and representative) sample. The resulting optimization problem is deterministic and a wide range of optimization methods with proven theoretical guarantees can be used.  For example, inexact Newton methods may be utilized to obtain fast convergence \cite{Bollapragada_2018,OLearyRoseberry:2019vf,Xu_2020}. Solving a deterministic SAA optimization problem with an efficient solver guarantees the linear model fits the sampled data optimally at each training iteration.  Note that if an SAA approach were used to solve both \cref{eqn:objFctnReducedphi} and \cref{eq:objFctnLinphi} with the same (fixed) sample set, then this would be equivalent to the variable projection SAA approach described in \cite{newman2020train}.  Indeed, there are various recent works~\cite{newman2020train, patel2020block, cyr2019robust} that exploit the separable structures~\eqref{eq:separableDNN} of neural networks in SAA settings in order to accelerate convergence.  However, the disadvantage of SAA methods is that very large batch sizes are needed to obtain sufficient accuracy of the approximation and to prevent overfitting. Although parallel computing tools (e.g., GPU and distributed computing) and strategies such as repeated sampling may be used, the storage requirements for SAA methods remain prohibitively large. 

To summarize~\cref{sec:separable}, the widely-used, fully-coupled approach (optimizing over $\bftheta$ and $\bfW$ simultaneously) and the alternating minimization approach represent two extremes: the former is a tractable approach, but ignores the separable structure while the latter exploits separability, but is computationally intractable in the stochastic setting.  Although a block coordinate descent approach decouples the parameters and replaces expensive optimization solves with iterative updates, a VarPro approach can mathematically eliminate the linear weights, thereby reducing the problem to a stochastic optimization problem in $\bftheta$ only.  The resulting noisy gradient estimates for $\bftheta$ are unbiased when $\widehat{\bfW}(\bftheta)$ is computed exactly, making VarPro compatible with SGD variants to update $\bftheta$.  However, computing $\widehat{\bfW}(\bftheta)$ when also updating $\bftheta$ is intractable and poor approximations may lead to a large bias in the gradients for $\bftheta$.  Hence, providing an effective and efficient way to approximate $\widehat{\bfW}(\bftheta)$ is crucial to obtain a practical implementation of VarPro stochastic optimization.

% ====================================================================== %
% Section 3 - slimTrain
% ====================================================================== %

\section{Sampled limited-memory DNN training with {\tt slimTrain}}
\label{sec:sa}

We present {\tt slimTrain} as a tractable variant of VarPro in the SA setting, which adopts a sampled limited-memory Tikhonov scheme to approximate the linear weights and to estimate an effective regularization parameter for the linear weights. The key idea is to approximate the linear weights using the output features obtained from recent mini-batches and nonlinear weight iterates. By storing the output features from the most recent iterates, {\tt slimTrain} avoids additional forward and backward propagations through the neural network which, especially for deep networks, is computationally the most expensive part of training, and hence adds only a small computational overhead to the training.

% --------------------------------------------------------------------------------------------------------------------------- %
\subsection{Sampled Tikhonov methods to approximate $\widehat{\bfW}(\bftheta)$}
\label{sub:slimTik}
As described in~\cref{sec:separable}, approximating $\widehat{\bfW}(\bftheta)$ well is challenging, but important for reducing bias in the gradient for $\bftheta$, see \cref{eq:unbiasedVarPro}. This motivates us to use state-of-the-art iterative sampling approaches to solve stochastic, Tikhonov-regularized, \textit{linear} least-squares problems.  For exposition purposes, we first reformulate \cref{eq:objFctnLinphi} as
\begin{equation}
\label{eq:linstochTik}
    \widehat \bfw(\bftheta) = \argmin_{\bfw} \ \Ebb \ \thf \left\| \bfA(\bfy, \bftheta) \bfw - \bfc\right\|_2^2 + \tfrac{\lambda}{2}\| \bfw\|_{\rm 2}^2,
\end{equation}
where $\bfw = {\rm vec}(\bfW) \in \Rbb^{\nTargets\nFeatOut}$ concatenates the columns of $\bfW$ in a single vector, $\bfA(\bfy, \bftheta) =  F(\bfy, \bftheta)^\top \otimes \bfI_{\nTargets}$ with $\otimes$ denoting the Kronecker product. This Kronecker structure extends to a mini-batch $\Tcal_k$.  Suppose we order the samples $(\bfy_i,\bfc_i) \in \Tcal_k$ for $i=1,\dots, |\Tcal_k|$.  Then, the final layer be expressed for vectorized linear weights as
    \begin{align*}
       \bfW \bfZ_k(\bftheta) \approx \bfC_k
       \qquad \begin{array}{c}
       \xrightarrow[\hspace{1cm}]{\text{vec}}\\[-1em] 
       \xleftarrow[\text{mat}]{\hspace{1cm}} \end{array} \qquad
      \bfA_k(\bftheta) \bfw \approx \bfb_k
    \end{align*}
where
    \begin{align*}
    	\bfZ_k(\bftheta)  &= \begin{bmatrix}F(\bfy_1,\bftheta) & \cdots & F(\bfy_{|\Tcal_k|},\bftheta) \end{bmatrix} &&\in \bbR^{\nFeatOut \times \batchSize},\\
	\bfC_k &= \begin{bmatrix} \bfc_1 & \cdots & \bfc_{\batchSize} \end{bmatrix} && \in \bbR^{\nTargets \times\batchSize},\\
        \bfA_k(\bftheta) &= \bfZ_k(\bftheta)^\top \otimes \bfI_{\nTargets} && \in \Rbb^{\batchSize\nTargets \times \nFeatOut \nTargets}, \quad\mbox{and}\\
        \bfb_k &= {\rm vec}(\bfC_k) = \begin{bmatrix} \bfc_1 \\ \vdots\\ \bfc_{\batchSize} \end{bmatrix} && \in \Rbb^{\nTargets \batchSize}.
    \end{align*}
Henceforth, in this section, since $\bftheta$ is fixed in~\cref{eq:linstochTik}, we use $\bfA_k = \bfA_k(\bftheta)$ for presentation purposes.

Introduced in~\cite{slagel2019sampled,chung2020sampled}, sampled Tikhonov ({\tt sTik}) and sampled limited-memory Tikhonov ({\tt slimTik}) methods are specialized iterative methods developed for solving stochastic regularized linear least-squares problems. For an initial iterate $\bfw_0$, the $k$-th {\tt sTik} iterate is given by
	\begin{equation}
	\label{eq:sTik}
    	\bfw_k(\Lambda) = \argmin_\bfw \frac{1}{2}\left\| \begin{bmatrix} \bfA_1 \\ \vdots \\ \bfA_{k-1} \\ \bfA_k \\ \sqrt{\Lambda + \sum_{i=1}^{k-1} \Lambda_i} \bfI
    	\end{bmatrix}
    	\bfw -  \begin{bmatrix} \bfA_1 \bfw_{k-1} \\ \vdots \\ \bfA_{k-1} \bfw_{k-1} \\ \bfb_k \\ \frac{\sum_{i=1}^{k-1} \Lambda_i}{\sqrt{\Lambda + \sum_{i=1}^{k-1} \Lambda_i}} \bfw_{k-1}
    	\end{bmatrix}\right\|_2^2,
	\end{equation}
where $\bfw_{k-1}$ is the previously computed estimate, $\bfA_1, \ldots, \bfA_{k-1}$ are matrices containing previously computed output features, $\Lambda +\sum_{i=1}^{k-1} \Lambda_i >0$, and $\Lambda$ is a regularization parameter estimate. The {\tt sTik} iterates can also be expressed in update form as an SA method,
	\begin{equation}
	\label{eq:samplediterative}
    	\bfw_k(\Lambda) = \bfw_{k-1} - \bfB_k(\Lambda) \bfg_k(\bfw_{k-1},\Lambda),
	\end{equation}
with $\bfg_k(\bfw_{k-1},\Lambda) = \bfA_k\t (\bfA_k \bfw_{k-1} - \bfb_k) + \Lambda \bfw_{k-1}$ containing gradient information for the current mini-batch and $\bfB_k(\Lambda) = ((\Lambda + \sum_{i=1}^{k-1} \Lambda_i) \bfI + \sum_{i=1}^k \bfA_i\t \bfA_i)^{-1}$ containing global curvature information of the least-squares problem. Note that contrary to standard SA methods, \cref{eq:samplediterative} does not require a learning rate nor a line search parameter. The learning rate can be interpreted as one, which is optimal for Newton's method. 

\begin{figure}[bt]
\includegraphics[width=\textwidth]{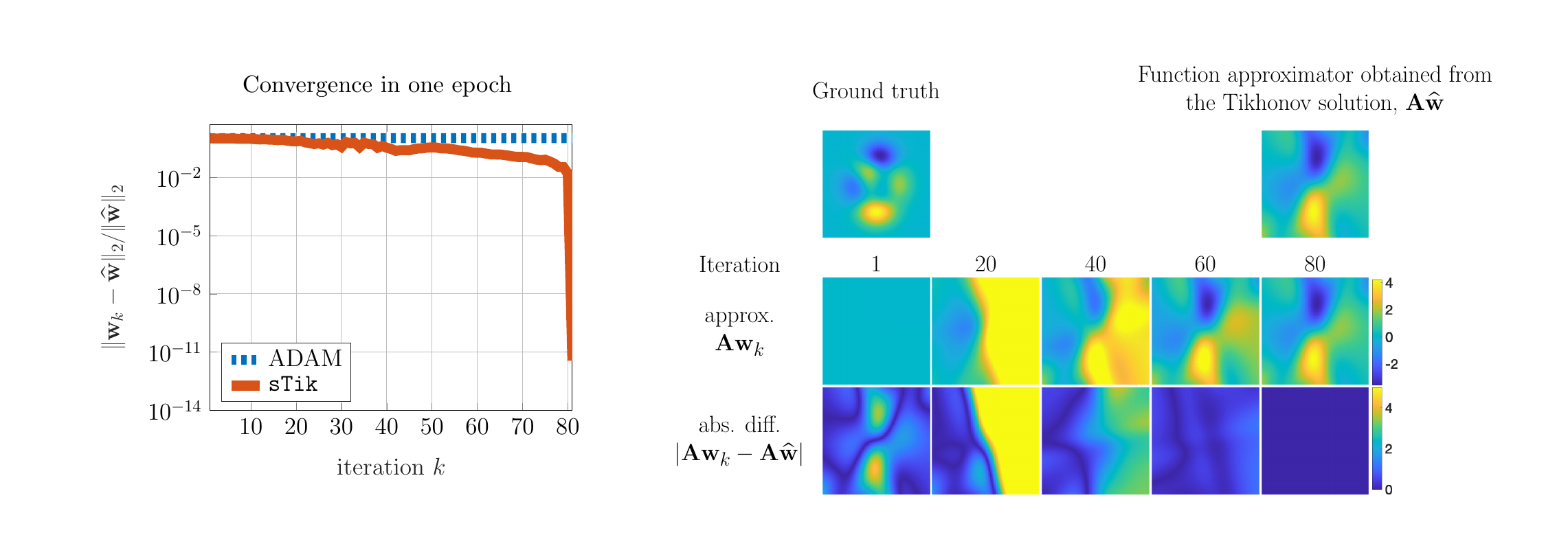}
\caption{
Illustration comparing convergence of {\tt sTik} and ADAM with a fixed regularization parameter for solving~\eqref{eq:linstochTik}. We consider approximating {\sc Matlab}'s \texttt{peaks} function, $f:\Rbb^2\to \Rbb$, using training data located on a uniform grid. We apply a fixed nonlinear transformation to each point in the domain to form the rows of $\bfA_k\in \Rbb^{|\Tcal_k|\times |\bfw|}$ and the corresponding true function values are stored in $\bfb_k\in \Rbb^{|\Tcal_k|}$ where $|\bfw|$ is the number of linear weights. The constant regularization parameters are $\Lambda_k = \frac{\lambda}{80}$ where $80$ is the number of iterations in one epoch. Both $|\bfw|$ and $\lambda$ are chosen arbitrarily and the number of iterations depends on the number of training points and the batch size. The best linear weights are given by the Tikhonov solution, $\widehat{\bfw} = (\bfA^\top \bfA +\lambda\bfI)^{-1}\bfA^\top \bfb$, and the corresponding best function approximator is $\bfA\widehat\bfw$. To the left, we plot the convergence of the relative error $\|\bfw_k - \widehat{\bfw}\|_2/\|\widehat{\bfw}\|_2$ for each iteration $k$ in a single epoch.  By design, \texttt{sTik} converges to the least-squares solution in one epoch whereas ADAM makes little progress.  To the right, the middle row shows the function approximations for different \texttt{sTik} iterates, $\bfA\bfw_k$, and the bottom row shows the absolute difference of the approximation with the best approximation.  The top row depicts the true \texttt{peaks} function (left) and the best approximation obtained from the Tikhonov solution (right). 
}
\label{fig:slimTikIntuition}
\end{figure}

Importantly, the regularization parameter $\lambda$ in \cref{eq:linstochTik}, which is typically required to be set in advance, has been replaced with a new parameter estimate $\Lambda$ which can be chosen adaptively at each iteration.  Each $\Lambda_k$ corresponds to a regularization parameter at iteration $k$ and can change at each iteration ($\Lambda_j$, $j = 1,\ldots,k-1$ correspond to regularization parameters from previous iterations). In fact, the parameters $\lambda$ and $\Lambda_k$'s  are directly connected. After one epoch (e.g., iterating through all training samples), the {\tt sTik} iterate is identical to the Tikhonov solution of \cref{eq:linstochTik} with $\lambda = \sum_{i=1}^k \Lambda_i$ where $k$ is the number of iterations required for one epoch.  We exemplify the convergence of {\tt sTik} in \cref{fig:slimTikIntuition} when approximating {\sc Matlab}'s \texttt{peaks} function \cite{HaberRuthotto2017}. Moreover, it has been shown that {\tt sTik} iterates converge asymptotically to a Tikhonov solution and subsequently adaptive parameter selection methods were developed in \cite{slagel2019sampled}. 

Since~\cref{eq:sTik} and~\cref{eq:slimTik} correspond to standard Tikhonov problems, extensions of standard regularization parameters methods, such as the discrepancy principle (DP), unbiased predictive risk minimization (UPRE), and generalized cross validation (GCV) techniques can be utilized. Indeed, \emph{sampled} regularization parameter selection methods sDP, sUPRE, and sGCV for {\tt sTik} and {\tt slimTik} and their connection to the overall regularization parameter $\lambda$ can be found in \cite{slagel2019sampled}. In this work, we focus on regularization parameter selection via sGCV since this method does not require any further hyperparameters (e.g., noise level estimates for the mini-batch), and we have observed that sGCV provides favorable $\lambda$ estimates.  For details on the GCV function, see original works~\cite{Golub1979:gcv,Wahba1977PracticalAS} and books~\cite{Hansen1998:rankDeficient,Vogel2002:inverseProblems}. The sGCV parameter at the $k$-th {\tt slimTik} iterate can be computed as
 \begin{equation}\label{eq:sGCV}
     \Lambda_k = \argmin_{\Lambda} \  \frac{|\calT_k| \ \|\bfA_k \bfw_k(\Lambda) - \bfb_k\|_2^2}{\left( |\calT_k| -\trace{\bfA_k \bfT_k(\Lambda) \bfA_k\t}\right)^2}
 \end{equation}
 where
 \begin{equation}
    \bfT_k(\Lambda) = \left(\left(\Lambda + \sum_{i=1}^{k-1} \Lambda_i \right)\bfI_n + \sum_{i=k-\memDepth}^k \bfA_i\t \bfA_i \right)^{-1}.
\end{equation}

For some problems, e.g., inverse problems where $\bfA_k$ represent large-scale forward model matrices, {\tt sTik} may not be practical since each iteration requires either solving a least-squares problem \cref{eq:sTik} whose coefficient matrix is growing at each iteration or updating matrix $\bfB_k$.  To alleviate the memory burden, a variant of {\tt sTik} called the sampled limited-memory Tikhonov ({\tt slimTik}) method was proposed in \cite{slagel2019sampled}. Let $\memDepth \in \bbN_0$ be a memory depth parameter. Then, the $k$-th {\tt slimTik} iterate has the form 
\begin{equation}
\label{eq:slimTik}
 \bfw_k (\Lambda) = \argmin_{\bfw} \frac{1}{2} \norm[2]{
    \begin{bmatrix}
    \bfA_{k-\memDepth} \\ \vdots \\ \bfA_{k-1} \\ \bfA_{k}\\ \sqrt{ \Lambda + \sum_{i = 1}^{k-1} \Lambda_i}\bfI
    \end{bmatrix}  \bfw -
     \begin{bmatrix}
     \bfA_{k-\memDepth} \bfw_{k-1} \\ \vdots \\ \bfA_{k-1} \bfw_{k-1} \\\bfb_k\\ \frac{\sum_{i = 1}^{k-1} \Lambda_i}{\sqrt{ \Lambda+\sum_{i = 1}^{k-1} \Lambda_i}}\bfw_{k-1}
    \end{bmatrix}}^2.
\end{equation}

We provide a few remarks about the {\tt slimTik} method.  For linear least-squares problems, it can be shown that for the case $r=0$, the {\tt slimTik} method is equivalent to the stochastic block Kaczmarz method.  Furthermore, for linear least-squares problems with a fixed regularization parameter, theoretical convergence results for {\tt slimTik} with memory $r=0$ were developed in \cite{chung2020sampled}. We point out that limited memory methods like {\tt slimTik} were initially developed to address problems where the size of $\bfw$ is massive, but this is not necessarily the case in DNN training where the number of weights in $\bfw$ may be modest. However, as we will see in \cref{sub:saslimtik}, a limited memory approach is suitable and can even be desirable in the context of solving nonlinear problems, where nonlinear parameters have direct impact on the model matrices $\bfA_k$.  In this work, we are interested in incorporating extensions of {\tt slimTik} with \textit{adaptive} regularization parameter selection for nonlinear problems that exploit separability. 

% --------------------------------------------------------------------------------------------------------------------------- %
\subsection{\texttt{slimTrain}}
\label{sub:saslimtik}

Our proposed SA algorithm, {\tt slimTrain} takes advantage of the separable structure of many DNNs and integrates the {\tt slimTik} method for efficiently updating the linear parameters and for automatic regularization parameter tuning.  We consider the {\tt slimTik} update of $\bfW$ to serve as an approximation of the eliminated linear weights in VarPro SA from~\eqref{eq:objFctnLinphi}. Specifically, at the $k$-th iteration, $\widehat \bfW(\bftheta) \approx \bfW_k = {\rm mat}(\bfw_k(\Lambda_k))$ where
	\begin{equation}
	\label{eq:slimTik_nonlin}
	 \bfw_k(\Lambda_k) = \argmin_{\bfw} \ \norm[2]{
    		\begin{bmatrix}
   	 	\bfM_{k} \\ \bfA_{k}(\bftheta_{k-1})\\ \sqrt{\sum_{i = 1}^{k} \Lambda_i}\bfI
   	 	\end{bmatrix}  \bfw -
    		 \begin{bmatrix}
     		\bfM_k \bfw_{k-1} \\\bfb_k\\ \frac{\sum_{i = 1}^{k-1} \Lambda_i}{\sqrt{\sum_{i = 1}^{k} \Lambda_i}}\bfw_{k-1}
    		\end{bmatrix}}^2,
	\end{equation}
with
\begin{equation}\label{eq:updateM}
  \bfM_k = 
  \begin{bmatrix}
      \bfA_{k-\memDepth}(\bftheta_{k-\memDepth-1}) \\ \vdots \\ \bfA_{k-1} (\bftheta_{k-2})
  \end{bmatrix}
\end{equation} 
and $\Lambda_k$ is computed using the sGCV method (c.f., \cref{eq:sGCV}). Notice that this is not equivalent to the {\tt slimTik} method for $\argmin_\bfW \Phi(\bfW, \bftheta_{k-1})$, since there is no inner iterative process and because of the dependence on previous $\bftheta_j$. A summary of the algorithm is provided in \cref{alg:decoupledSA}.

\begin{algorithm}
    \begin{algorithmic}[1]
    \State \textbf{Training Data:} $\Tcal\subseteq \Dcal$
    \State \textbf{Hyperparameters:} memory depth $\memDepth \in \bbN_0$, mini-batch size $n_{\rm batch}$, learning rate $\gamma$, regularization parameter $\alpha$
    \State \textbf{Initialize:} $\bftheta_0 \in \bbR^{\nTheta}$, $\bfW_0\in \bbR^{\nTargets \times \nFeatOut}$
    \medskip
    \While{not converged} % iterate over epochs
    	\State randomly partition $\Tcal$ into mini-batches such that $\Tcal=\bigsqcup_k \Tcal_k$ and $|\Tcal_k| = n_{\rm batch}$

        \For{$k=1,\dots,\lfloor |\Tcal| / n_{\rm batch}\rfloor$} % describes one epoch
        
        \State select mini-batch $\Tcal_k$ 
        \State forward propagate network to obtain $\bfA_k(\bftheta_{k-1})$ 
        \State update memory matrix $\bfM_k$ \algorithmiccomment{\cref{eq:updateM}}
        \State select $\Lambda_k$ using sGCV \algorithmiccomment{\cref{eq:sGCV}}
        \State compute $\bfW_k = {\rm mat}(\bfw_k(\Lambda_k))$  \algorithmiccomment{\cref{eq:slimTik_nonlin}}
        \State compute $\left[{\rm D}_\bftheta \Phi_k(\bfW_k,\bftheta)\right]_{\bftheta = \bftheta_{k-1}}$ via backpropagation \algorithmiccomment{\cref{eq:batchGradExplicit}}
        \State select search direction $\bfp_k$ 
        \State update $\bftheta_k = \bftheta_{k-1} + \gamma \bfp_k(\bftheta_{k-1};\bfW_k)$
        \EndFor
    \EndWhile
    \end{algorithmic}
    \caption{{\tt slimTrain}: sampled limited-memory training for separable DNNs}
    
    \label{alg:decoupledSA}
\end{algorithm}

We note that an SA method that incorporates the {\tt slimTik} method was considered for separable nonlinear inverse problems in \cite{chung2019iterative}, but there are some distinctions.  First, the results in \cite{chung2019iterative} use a fixed regularization parameter, but here we allow for adaptive parameter choice, which has previously only been considered for linear problems.  We note that updating regularization parameters in nonlinear problems (especially stochastic ones) is a challenging task, and currently there are no theoretical justifications.  Second, all forward matrices were recomputed for each new set of nonlinear parameters in \cite{chung2019iterative}.  That is, for updated estimate $\bftheta_{k-1},$
\begin{equation}
  \bfM_k = 
  \begin{bmatrix}
      \bfA_{k-\memDepth}(\bftheta_{k-1}) \\ \vdots \\ \bfA_{k-1}(\bftheta_{k-1})
  \end{bmatrix}.
\end{equation}
Such an approach would be computationally demanding for DNN learning problems, since this would require revisiting previous mini-batches and re-computing the forward propagation matrix for new parameters $\bftheta_{k-1}$. Instead, we propose to use \cref{eq:updateM}, and we will show that these methods can perform well in practice.

% --------------------------------------------------------------------------------------------------------------------------- %
\subsection{Efficient implementation}
\label{sec:implementation}

Training separable DNNs with {\tt slimTrain} \linebreak adds some computational costs compared to existing SA methods like ADAM; however, those are modest in many cases and the overhead in computational time can be reduced by an efficient implementation.  The additional costs stem from  solving for the optimal linear weights in~\eqref{eq:slimTik} and approximating the optimal regularization parameter using the sGCV function~\eqref{eq:sGCV}.  The costs of these steps depend on the size of the nonlinear feature matrix, $\bfA_k \in \Rbb^{\batchSize\nTargets \times \nFeatOut\nTargets}$, the size of the memory matrix, $\bfM_k$, which contains $r$ blocks of nonlinear features from previous batches, and the number of linear weights. In the case when the linear weights are applied via dense matrix, we can exploit the Kronecker structure in our problem; see~\cref{sub:slimTik} for details.  The Kronecker structure results in solving $\nTargets$ least-squares problems simultaneously where each problem is moderate in size (typically, on the order of $10^2$ or $10^3$).   Due to the modest problem size, we use a singular value decomposition (SVD) to solve the least-squares problem. We also re-use the SVD factors for efficiently adapting the regularization parameter.  For the \texttt{peaks} and surrogate modeling experiments (\cref{sec:peaks} and~\cref{sec:surrogate}), we implement the Kronecker-structure framework in {\sc Matlab}.   The code is available in the {\tt Meganet.m} repository on \url{https://github.com/XtractOpen/Meganet.m}.  

In the case when the linear weights parameterize a linear operator (most importantly, a convolution), efficient iterative solvers, such as LSQR~\cite{Paige82lsqr:an} that only require matrix-vector products and avoid forming the matrix explicitly, can be used to find the optimal linear weights. Such methods were employed in~\cite{chung2019iterative} where the authors applied {\tt slimTik} to massive, separable nonlinear inverse problems where the data matrix could not be represented all-at-once.  Modifications of the sGCV function using stochastic trace estimators can then be used for estimating the regularization parameter efficiently; for more details, see~\cite{slagel2019sampled}.

In~\cref{sec:autoencoder}, the linear weights parameterize a convolution layer with several input but only one output channel. Exploiting the separability between the different channels and the small number of weights per channel, we form the nonlinear feature matrix, $\bfA_k$, explicitly in our implementation.  This allows us to use the same SVD-based automatic regularization parameter selection as in the dense case. To be precise, the columns of $\bfA_k$ are shifted copies of the batch data, which is large, but accessible (on the order of $10^5$).  Importantly, the number of columns (copies of the data) is small because the number of weights parameterizing the linear operator, denoted $|\bfw|$, is small (on the order of $10^2$).  We can construct the data matrix $\bfA_k$ efficiently by taking advantage of the structure of convolutional operators; each channel has its own linear weights and the samples share the same weights. For storage efficiency, we can form the smaller matrix $\bfA_k^\top \bfA_k \in \Rbb^{|\bfw|\times |\bfw|}$ one time, and use the update rule~\eqref{eq:samplediterative} to adjust the linear weights. We implement the convolutional operator framework in Pytorch~\cite{pytorch}. The code is available on {\tt github} at \url{https://github.com/elizabethnewman/slimTrain}.

% ====================================================================== %
% Section 4 - slimTrain
% ====================================================================== %

\section{Numerical results}
\label{sec:results}

We present a numerical study of training separable DNNs using {\tt slimTrain} with automatic regularization parameter selection. In this section, we first provide a general discussion on numerical considerations of our proposed method in \cref{sec:implementation}. In~\cref{sec:peaks}, we explore the relationship between various {\tt slimTrain} hyperparameters (e.g., batch size, memory depth, regularization parameters) in a function approximation task.  Our results show that automatic regularization parameter selection can mitigate poor hyperparameter selection. In~\cref{sec:surrogate}, we apply {\tt slimTrain} to a PDE surrogate modeling task and show that it outperforms the state-of-the-art ADAM for the default hyperparameters.  In~\cref{sec:autoencoder}, we apply {\tt slimTrain} to a dimensionality-reduction task in which the linear weights are applied via a convolution. Notably, we observe faster convergence and, particularly with limited training data, improved results compared to ADAM.

% --------------------------------------------------------------------------------------------------------------------------- %
\subsection{Peaks}
\label{sec:peaks}

\begin{figure}
\centering
\includegraphics[width=\textwidth]{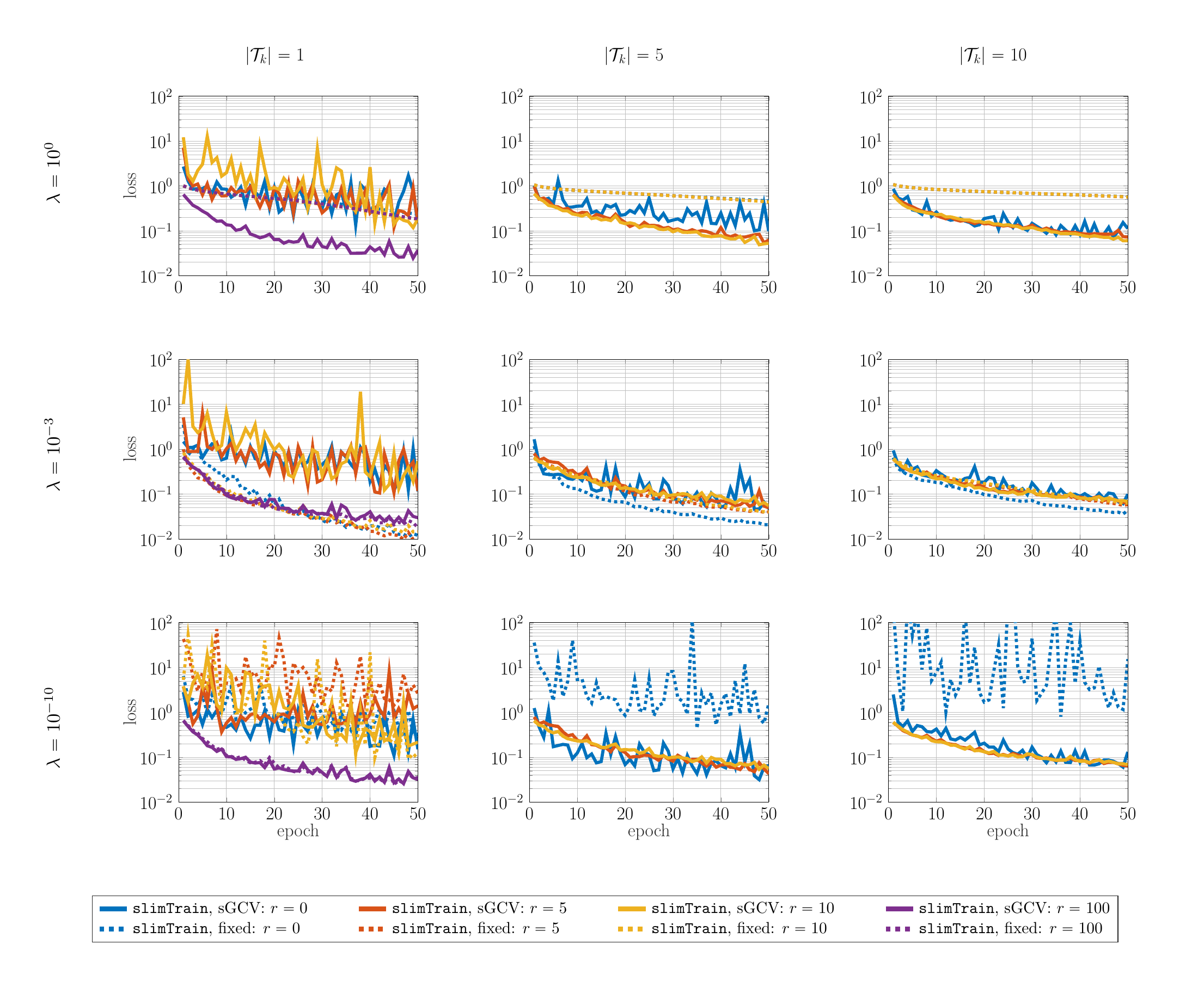}

\caption{Convergence of training loss for the \texttt{peaks} experiment when training with {\tt slimTrain} with a learning rate of $\gamma=10^{-3}$. 
Each row corresponds to a different choice of fixed regularization parameter for $\bfW$, $\lambda=10^0, 10^{-3}, 10^{-10}$.  When training with adaptive regularization parameter selection, the initial regularization parameter $\Lambda_0$ is set to be the same as the fixed regularization parameter.  Each column corresponds to a different batch size, $|\Tcal_k|=1, 5, 10$. Each convergence plot consists of dashed and solid lines corresponding using a fixed regularization parameter and adaptively choosing the regularization parameter using sGCV, respectively. The color of each line corresponds to memory depth $r=0,5,10$ and, additionally, $r=100$ for $|\Tcal_k|=1$.}
\label{fig:peaksConvergenceStudy}
\end{figure}

To explore the hyperparameters in {\tt slimTrain}, we examine a scalar function approximation task. We train a DNN to fit the \texttt{peaks} function in {\sc Matlab},  which is a mixture of two-dimensional Gaussians. We use a small residual neural network (ResNet)~\cite{he2016deep} with a width of $w=8$ and a depth of $d=8$ corresponding to a final time of $T=5$.  Further details about the ResNet architecture can be found in~\cref{app:resnet}. The nonlinear feature extractor maps $F:\Rbb^2\times \Rbb^{528} \to \Rbb^8$ where $528$ is the number of weights in $\bftheta$.  The final linear layer introduces the weights $\bfW\in \Rbb^{1 \times 9}$, where the number of columns equals the width of the ResNet plus an additive bias. Our training data consists of 2,000 points sampled uniformly on the domain $[-3,3]\times [-3,3]$. We display the convergence of {\tt slimTrain} for various combinations of hyperparameters in~\cref{fig:peaksConvergenceStudy}. 

The interplay between number of output features, the batch size, and the memory depth is apparent in~\cref{fig:peaksConvergenceStudy}.  In this scalar-function example, we seek $9$ weights (i.e., $\bfW\in \Rbb^{1 \times 9}$) to fit $(r + 1)|\Tcal_k|$ samples. With small memory depth and batch size, the problem is underdetermined (or not sufficiently overdetermined) and solving for $\bfW$ significantly overfits the given batch at each iteration. This results in the slow, oscillatory convergence behavior, particularly with a batch size of $|\Tcal_k|=1$ (\cref{fig:peaksConvergenceStudy}, first column). When the memory depth and batch size are large enough (e.g., $r=100$ in the $|\Tcal_k|= 1$), the linear least-squares problem is sufficiently overdetermined and the training loss converges faster and to a lower value (\cref{fig:peaksConvergenceStudy}, purple line in first column).

\begin{figure}
\centering

\includegraphics[width=\textwidth]{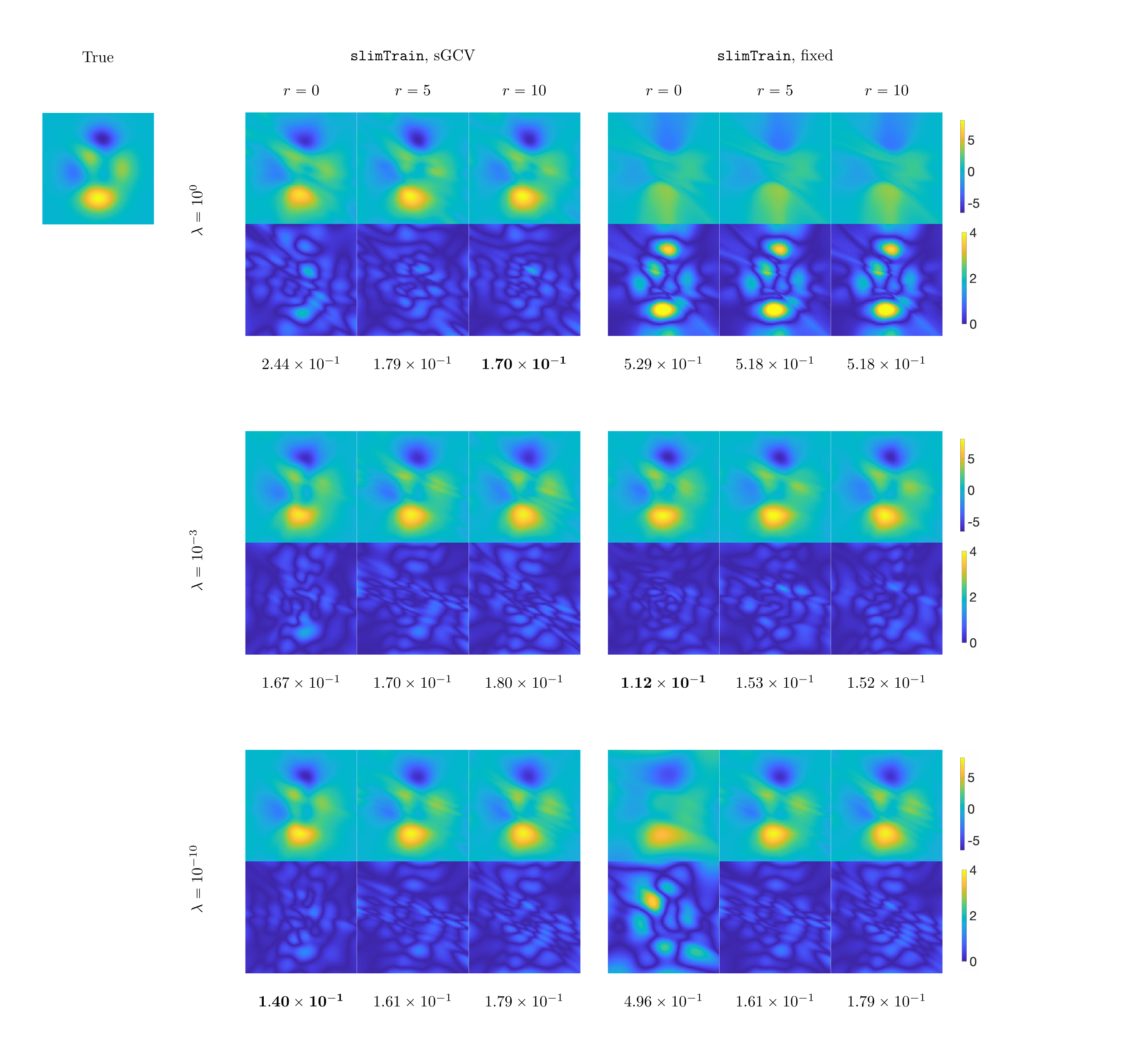}

\caption{DNN approximations for the \texttt{peaks} experiment with batch size of $|\Tcal_k|=5$ and a learning rate of $\gamma=10^{-3}$.  The results use the network weights corresponding to the lowest validation loss for each training method. Each block row corresponds to a different choice of fixed regularization parameter for $\bfW$, $\lambda=10^0, 10^{-3}, 10^{-10}$. The top rows of images in each block depict the DNN approximations of the \texttt{peaks} function.  The bottom rows of images in each block depict the absolute difference of the DNN approximations and the true \texttt{peaks} function. The DNN weights used provided the smallest validation loss during training. The relative error of the DNN approximation versus the true function is displayed below the corresponding absolute difference image.}
\label{fig:peaksApproxStudy}
\end{figure}

Solving the optimization problem and decreasing the loss of the training data is a proxy to the goal of DNN training: to generalize to unseen data.  To illustrate the generalizability of DNNs trained with {\tt slimTrain}, we display the DNN approximations in~\cref{fig:peaksApproxStudy} corresponding to a batch size of $|\Tcal_k|=5$ (second column of \cref{fig:peaksConvergenceStudy}) of the convergence plots.  

Exemplified in~\cref{fig:peaksApproxStudy}, the choice of regularization parameter for $\bfW$ significantly impacts the approximation quality of the network when training with a fixed regularization parameter (\cref{fig:peaksApproxStudy}, second column set of figures). If the optimization problem over-regularizes the linear weights ($\lambda=10^0$), the DNN approximation is smoother than the true \texttt{peaks} function and does not fit the extremes tightly (\cref{fig:peaksApproxStudy}, first row). In the under-regularized case ($\lambda=10^{-10}$) with a small memory depth ($r=0$), $\bfW$ overfits the batches and the DNN approximation does not generalize well (e.g., we miss the small peaks) (\cref{fig:peaksApproxStudy}, third row).  With a well-chosen regularization parameter (here, $\lambda=10^{-3}$), the DNN approximation is close to the true \texttt{peaks} function, but tuning this regularization parameter can be costly (\cref{fig:peaksApproxStudy}, second row). In comparison, the DNN approximations when automatically choosing a regularization parameter using the sGCV method are good approximations and look similar, no matter the initial regularization parameter or memory depth (\cref{fig:peaksApproxStudy}, first column set of figures). 

\begin{figure}
\centering
\includegraphics[width=0.8\textwidth]{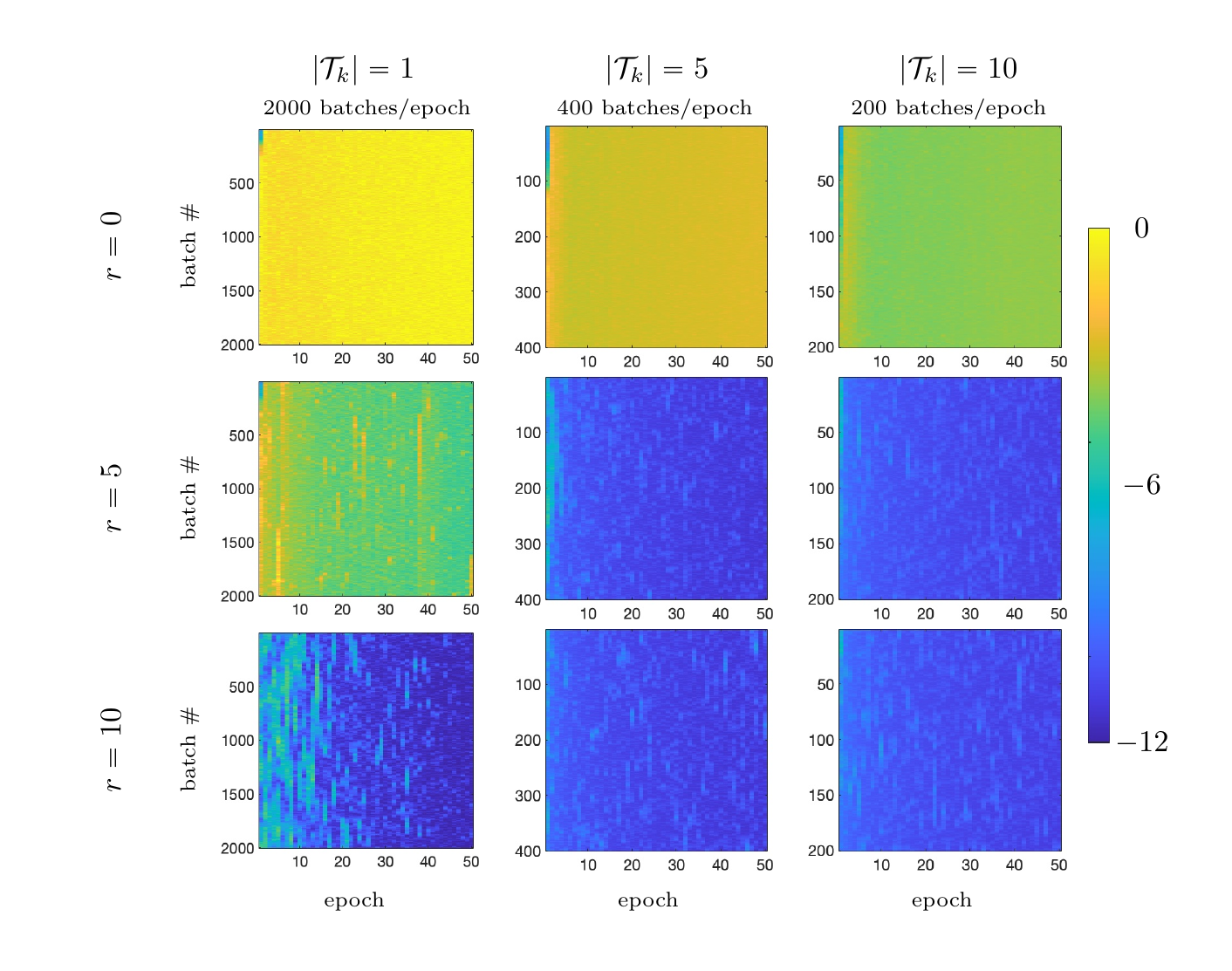}
\caption{Regularization parameters selected by approximately minimizing the sGCV function in the \texttt{peaks} example for a learning rate of $\gamma=10^{-3}$ and an initial regularization parameter of $\Lambda_0=10^{-10}$. Each column corresponds to a different batch size, $|\Tcal_k|=1,5,10$, respectively. Each row corresponds to a different memory depth, $r=0,5,10$, respectively. In each image, the horizontal axis is the number of epochs, in this case $50$, and the vertical axis is the number of iterations per epoch. For example, when the batch size is $|\Tcal_k|=5$, the vertical axis has $400$ iterations (the number of training samples divided by the batch size).  Each pixel corresponds to the regularization parameter used for a particular batch batch and the batches change because we shuffle the training data at the start of each epoch. The images are displayed in log scale. The first few regularization parameters in each case are small (top left corner of each image) because we start with a small initial regularization parameter.}
\label{fig:peaksAlphaStudy}
\end{figure}

The selected regularization parameters are related to the ill-posedness of the problem, as illustrated for the $\lambda=10^{-10}$ case in~\cref{fig:peaksAlphaStudy}. When the batch size is $|\Tcal_k|=1$ (\cref{fig:peaksAlphaStudy}, first column), the linear least-squares problem is underdetermined for memory depths $r=0$ and $r=5$ and is overdetermined when $r=10$.  To avoid overfitting in the underdetermined cases, larger regularization parameters are selected. In the overdetermined case, overfitting is less likely and thus less regularization is needed.

With an adequate choice of memory depth and batch size, training a DNN with {\tt slimTrain} decreases the training loss and generalizes well to unseen data. The choice of regularization parameter significantly impacts the resulting network: too much regularization and the training stagnates; too little regularization and the training oscillates. Employing adaptive regularization parameter selection mitigates these extremes and simplifies the costly a priori step of tuning the parameter.

% --------------------------------------------------------------------------------------------------------------------------- %
\subsection{PDE surrogate modeling}
\label{sec:surrogate}

Due to their approximation properties, there has been increasing interest in using DNNs as efficient surrogate models for computationally expensive tasks arising in scientific applications. One common task is partial differential equation (PDE) surrogate modeling in which a DNN replaces expensive linear system solves~\cite{olearyroseberry2021derivativeinformed, bhattacharya2021model, Zhu_2019,Tripathy:2018kk}. Here, we consider a parameterized PDE
    \begin{align}
        \bfc = \Pcal u \quad \text{where} \quad \Acal(u, \bfy) = 0,
    \end{align}
where $u$ is the solution to a PDE defined by $\Acal$ and parameterized by $\bfy$ (which could be discrete or continuous).  In our case, the solution is measured at discrete points given by the linear operator $\Pcal$ and the observations are contained in $\bfc$. The goal is to train a DNN as a surrogate mapping from parameters $\bfy$ to observables $\bfc$ and avoid costly PDE solves. 

\begin{figure}
\centering
\includegraphics[width=\textwidth]{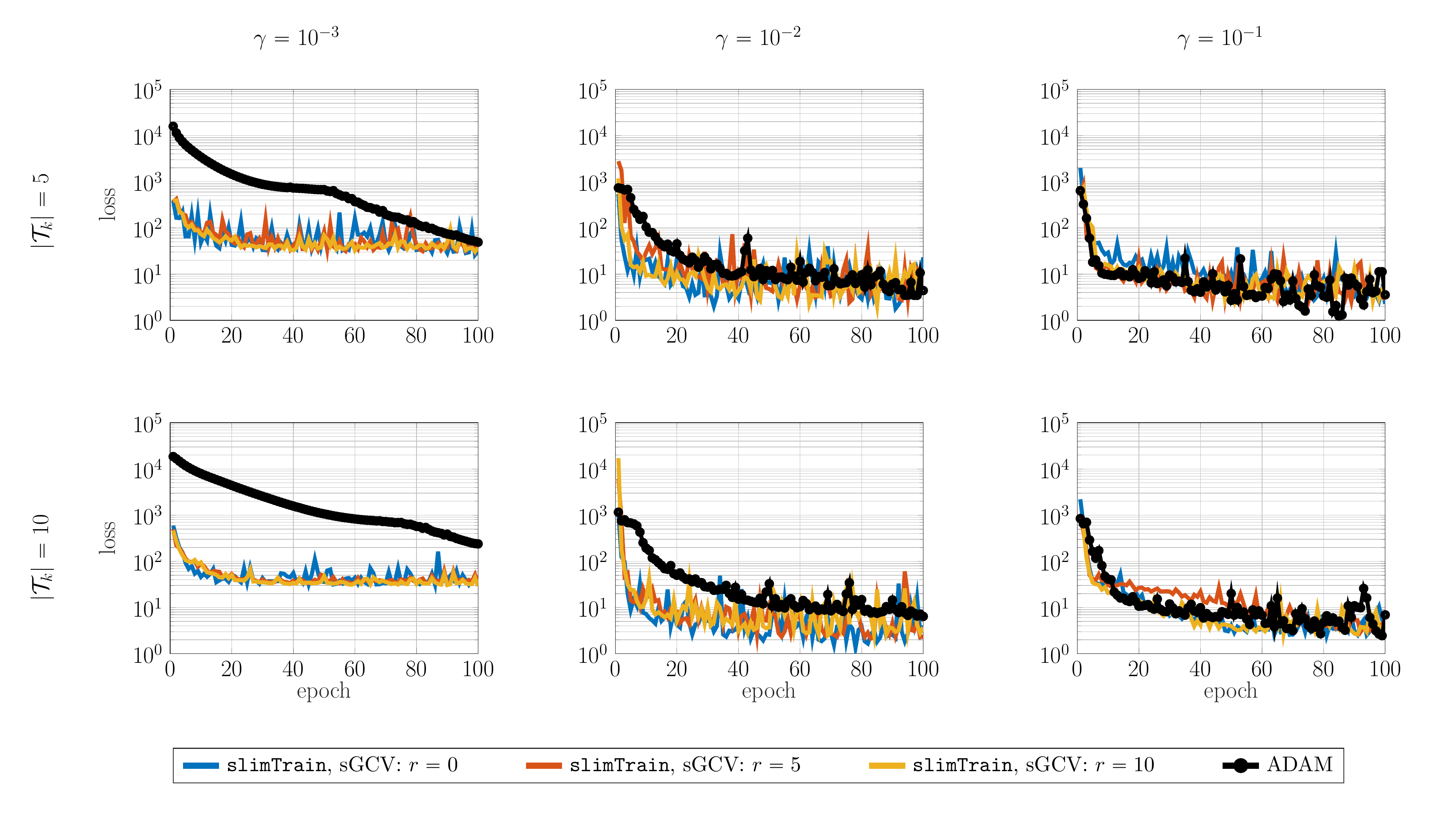}

\caption{Convergence results for the training loss for the CDR experiment. The rows correspond to different batch sizes, $|\Tcal_k| = 5,10$, and the columns correspond to different learning rates, $\gamma=10^{-3}, 10^{-2},10^{-1}$. The colorful, solid lines depict the convergence of the training loss using {\tt slimTrain} with sGCV regularization parameter selection.  Each color corresponds to a different memory depth, $r=0,5,10$. The black line with markers depicts the convergence of the training loss using ADAM.}
\label{fig:cdrParameterStudy}
\end{figure}

In our experiment, we consider the convection diffusion reaction (CDR) equation which models physical phenomena in many fields including climate modeling~\cite{stocker2011introduction} and mathematical biology~\cite{deVries2006:mathBio, Britton1986:cdrBiology}. As its name suggests, the CDR equation is composed of three terms: a diffusion term that encourages an even distribution of the solution $u$ (e.g., chemical concentration), a convection (or advection) term that describes how the flow (e.g., of the fluid containing the chemical) moves the concentration, and a reaction term that captures external factors that affect the concentration levels. In our example, the reaction term is a linear combination of $55$ different reaction functions and the parameters $\bfy\in \Rbb^{55}$ are the coefficients. The observables $\bfc\in \Rbb^{72}$ are measured at the same $6$ spatial coordinates and $12$ different time points; for details, see~\cite{newman2020train}.  We train a ResNet with a width of $w=16$ and a depth of $d=8$ corresponding to a final time of $T=4$; see~\cref{app:resnet} for further details. The linear weights in the final, separable layer are stored as a matrix $\bfW \in \Rbb^{72\times 17}$, where the number of columns is the width of the ResNet plus an additive bias. The results of training the ResNet with {\tt slimTrain} are displayed in~\cref{fig:cdrParameterStudy}. The major takeaway is that {\tt slimTrain} exploits the separable structure of the ResNet and, as a result, trains the network faster and fits the observed data better (lower loss) than ADAM with the recommended learning rate ($\gamma=10^{-3}$).

\begin{table}
\setlength\extrarowheight{5pt}

    \centering
    \caption{Training and validation loss in the CDR experiment for batch size $|\Tcal_k|=5,10$ and learning rates $\gamma=10^{-3}, 10^{-2}, 10^{-1}$. We display the loss after the first $20$ epochs to compare early performance. Because the memory depth does not significantly impact convergence, we display the loss for {\tt slimTrain} with a memory depth of $r=0$. Closeness between the training and validation losses indicates good generalization. The best overall performance (lowest loss) is achieved by {\tt slimTrain} with a batch size of $|\Tcal_k|=10$, denoted in bold.}
    \label{tab:cdrValidation}
    
\small
\begin{tabular}{|c|c||rr|rr|rr|}
\hline
  \multicolumn{1}{|c}{} && \multicolumn{2}{c|}{$\gamma=10^{-3}$} & \multicolumn{2}{c|}{$\gamma=10^{-2}$} & \multicolumn{2}{c|}{$\gamma=10^{-1}$}\\
   \cline{3-8}
 \multicolumn{1}{|c}{} & & \multicolumn{1}{c}{train} & \multicolumn{1}{c|}{valid}
    & \multicolumn{1}{c}{train} & \multicolumn{1}{c|}{valid}
    & \multicolumn{1}{c}{train} & \multicolumn{1}{c|}{valid}\\
\hline\hline
\multirow{2}{*}{\rotatebox{90}{  \scriptsize $|\Tcal_k|=5$~~}} 
& {\tt slimTrain}, $r=0$
    & $42.98$ & $41.17$
    & $22.06$ & $22.25$
    & $18.74$ & $23.25$ \\
& ADAM
    & $1453.00$ & $1338.00$
    & $45.24$ & $42.73$ 
    & $8.07$ & $8.70$ \\[1ex]
\hline
\multirow{2}{*}{\rotatebox{90}{\scriptsize $|\Tcal_k|=10$~~}} & 
  {\tt slimTrain}, $r=0$
    & $47.65$ & $52.95$
    & $\bf 4.28$ & $\bf 5.30$
    & $15.61$ & $16.60$ \\
& ADAM
    & $4405.00$ & $4143$
    & $49.92$ & $41.23$
    & $10.67$ & $10.71$\\[2ex]
\hline
\end{tabular}

\end{table}

In~\cref{tab:cdrValidation}, we examine if the performance of {\tt slimTrain} and ADAM generalizes to unseen after $20$ epochs; we choose $20$ epochs to analyze early performance and because the training loss decreases more closes after $20$ epochs in~\cref{fig:cdrParameterStudy}. The training and validation losses are close for both {\tt slimTrain} and ADAM, indicating that both training algorithms produce networks that generalize well.  For ADAM's suggested learning rate, $\gamma=10^{-3}$, {\tt slimTrain} achieves a validation loss that is two orders of magnitude less than that of ADAM.  When the learning rate is tuned to $\gamma=10^{-1}$, the performance of ADAM improves, but the overall best performance is achieved by {\tt slimTrain}. Most significantly, the performance of {\tt slimTrain} is less sensitive to the choice learning rate.

\begin{figure}
\centering

\includegraphics[width=0.8\textwidth]{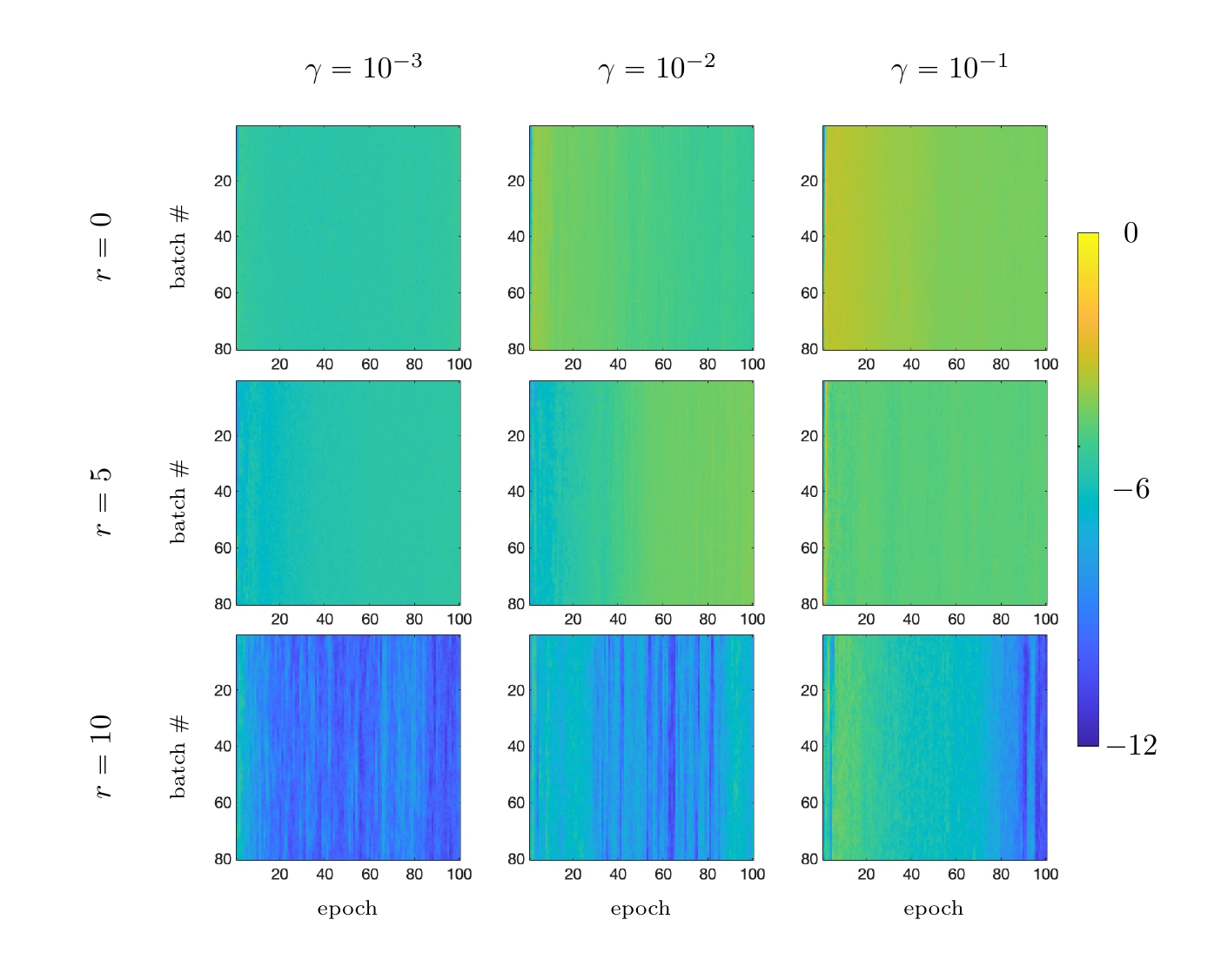}

\caption{Effect of learning rate and memory depth on the choice of regularization parameters in the CDR experiment.  The presented plots are from training using {\tt slimTrain} for a batch size of $|\Tcal_k| = 5$.  We show the regularization parameters (in log scale) obtained for various learning rates (columns) and memory depths (rows).}
\label{fig:cdrRegularization}

\end{figure}

As with the numerical experiment in~\cref{sec:peaks}, there is a relationship between batch size, memory depth, and the number of output features. In this experiment, because $\bfW\in \Rbb^{72\times 17}$, we solve $72$ independent least-squares problems with $17$ unknowns in each problem.  Illustrated in~\cref{fig:cdrRegularization}, when the memory depth is small ($r=0,5$), each least-squares problem is underdetermined or not sufficiently overdetermined, and hence more regularization on $\bfW$ is needed to avoid overfitting.  Because we use sGCV to automatically select the regularization parameter, the training with {\tt slimTrain} achieves a comparable loss for all memory depths.  In addition, the learning rate to update $\bftheta$ plays a role in the regularization parameters chosen.  When the learning rate is large ($\gamma=10^{-1}$), the output features of the network can change rapidly.  As a result, larger regularization parameters are selected, even in the sufficiently overdetermined case ($r=10$), to avoid fitting features that will change significantly at the next iteration.

In this surrogate modeling example, {\tt slimTrain} converges faster to the same or a better accuracy than ADAM using the recommended learning rate ($\gamma=10^{-3}$) by exploiting the separability of the DNN architecture.  Tuning the learning rate can improve the results for ADAM, but training with {\tt slimTrain} produces comparable results and reaches a desirable loss in the same or fewer epochs.  Using sGCV to select the regularization parameter on the weights $\bfW$ provides more robust training, adjusting automatically to the various hyperparameters (memory depth, learning rate) to produce consistent convergence.

% --------------------------------------------------------------------------------------------------------------------------- %
\subsection{Autoencoders}
\label{sec:autoencoder}

\begin{figure}
	\centering

	\includegraphics[width=\textwidth]{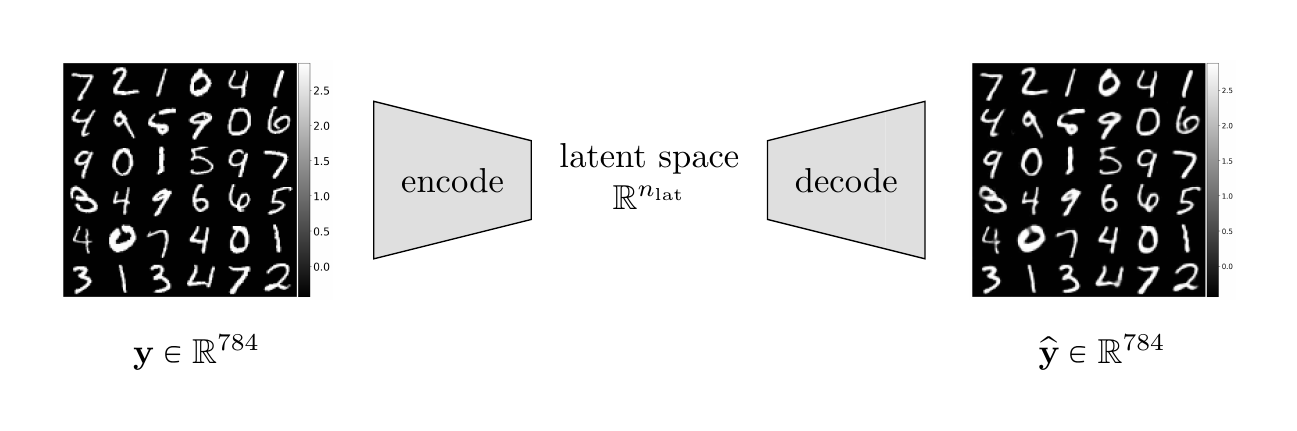}

	\caption{Illustration of autoencoder for the MNIST data.  The goal is to represent high-dimensional data in a low-dimensional, latent space for dimensionality reduction and feature extraction~\cite{Goodfellow:2016wc}. The encoder is a neural network that maps input data $\bfy$ to the latent space with intrinsic dimension ${n_{\rm lat}}$ (typically user-defined).  The decoder is a neural network that maps from the latent space to obtain an approximation of the original input, $\widehat{\bfy}$.}	
	\label{fig:autoencoderIllustration}

\end{figure}

Autoencoders are a dimensionality-reduction technique \linebreak using two neural networks: an encoder that represents high-dimensional data in a low-dimensional space and a decoder that reconstructs the high-dimensional data from this encoding, illustrated in \cref{fig:autoencoderIllustration}. Training an autoencoder is an unsupervised learning problem that can be phrased as optimization problem
	\begin{align}\label{eq:autoencoderObjFctn}
	\min_{\bfw,\bftheta_{\rm dec}, \bftheta_{\rm enc}} \Phi_{\rm auto}(\bfw,\bftheta_{\rm dec},\bftheta_{\rm enc}) &\equiv 
	\Ebb \ \tfrac{1}{2}\|\bfK(\bfw)F_{\rm dec}(F_{\rm enc}(\bfy,\bftheta_{\rm enc}),\bftheta_{\rm dec}) - \bfy\|_2^2 \\
	&\quad
		+ \tfrac{\alpha_{\rm enc}}{2} \|\bftheta_{\rm enc}\|_2^2
		+  \tfrac{\alpha_{\rm dec}}{2} \|\bftheta_{\rm dec}\|_2^2
		+ \tfrac{\lambda}{2} \|\bfw\|_2^2,\nonumber
	\end{align}
where the components of the objective function are the following:
	\begin{itemize}
	\item \textbf{Encoder:} $F_{\rm enc}: \Ycal \times \Rbb^{|\bftheta_{\rm enc}|} \to \Rbb^{n_{\rm lat}}$ is the encoding neural network that reduces the dimensionality of the input features $\nFeatIn$ to an \emph{intrinsic dimension} $n_{\rm lat}$ with $n_{\rm lat} \ll \nFeatIn$. Typically, the true intrinsic dimension is not known and must be chosen manually. The weights are $\bftheta_{\rm enc} \in \Rbb^{|\bftheta_{\rm enc}|}$, the number of encoder weights is $|\bftheta_{\rm enc}|$, and the regularization parameter is $\alpha_{\rm enc}\geq 0$. 
	
	\item \textbf{Decoder Feature Extractor:} $F_{\rm dec}: \Rbb^{n_{\rm lat}} \times \Rbb^{|\bftheta_{\rm dec}|} \to \Rbb^{\nFeatOut}$ is the decoder feature extractor. The weights are $\bftheta_{\rm dec} \in \Rbb^{|\bftheta_{\rm dec}|}$, the number of weights is $|\bftheta_{\rm dec}|$, and the regularization parameter is $\alpha_{\rm dec}\geq 0$. 
	
	\item \textbf{Decoder Final Layer:} $\bfK(\cdot): \Rbb^{|\bfw|} \to \Rbb^{\nFeatIn\times \nFeatOut}$ is a linear operator, mapping $\bfw$ to a matrix $\bfK(\bfw)$. For instance, $\bfK(\bfw)$ could be a sparse convolution matrix which can be accessed via function calls. The learnable weights $\bfw$ have a regularization parameter $\lambda\geq 0$.

	\end{itemize}
	
For notational simplicity, we let $\bftheta = (\bftheta_{\rm enc},\bftheta_{\rm dec})$ and $\alpha = \alpha_{\rm enc} = \alpha_{\rm dec}$ for the remainder of this section.	
	
In this experiment, we train a small autoencoder on the MNIST dataset~\cite{Lecun:1990mnist}. The data consists of 60,000 training and 10,000 test gray-scale images of size $28\times 28$ (i.e., $784$ input features). We implement convolutional neural networks for both the encoder and decoder with intrinsic dimension $n_{\rm lat}=50$; see details in~\cref{app:autoencoder}. Unlike the dense matrices in the previous experiments, the final, separable layer is a (transposed) convolution. Because convolutions use few weights and the prediction is high-dimensional, the least-squares problem is always overdetermined for this application. Hence, we require only a moderate memory depth in our experiments and, motivated by our results in~\cref{sec:peaks} and~\cref{sec:surrogate}, we use a memory depth of $r=5$ when training with {\tt slimTrain}. 

\begin{figure}
\centering

\begin{tabular}{c}
\includegraphics[width=0.9\textwidth]{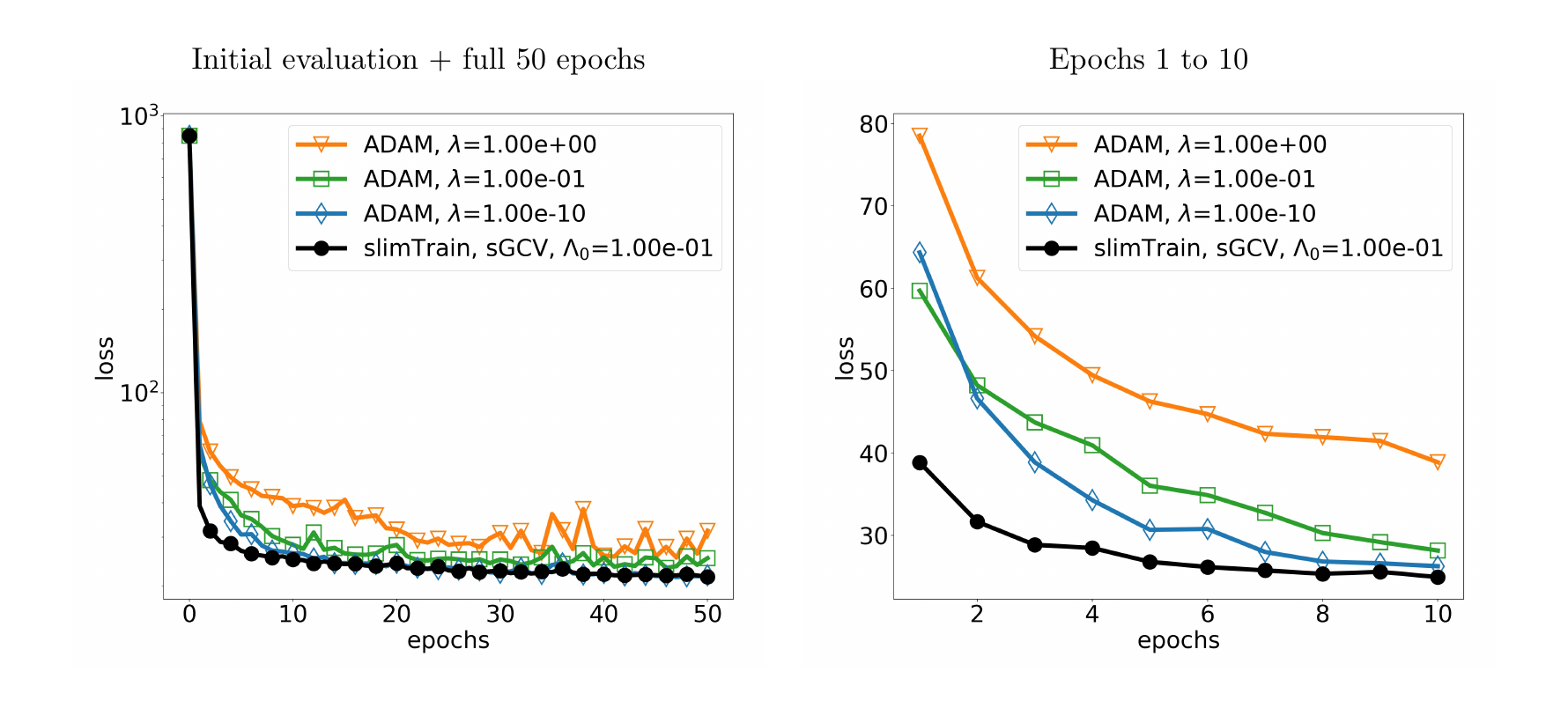}\\
DNN approximations after $1$ epoch\\
\includegraphics[width=0.9\textwidth]{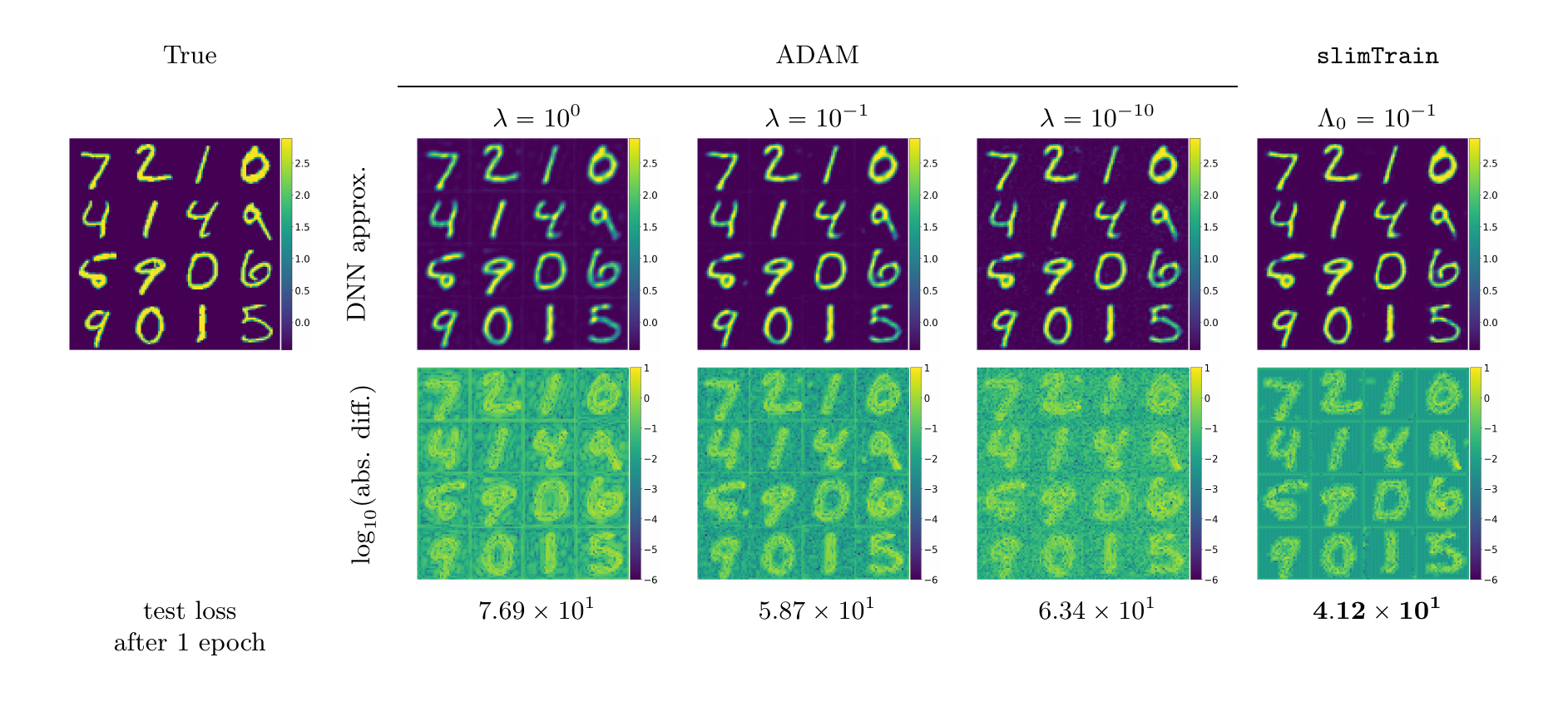}
\end{tabular}

\caption{Training loss convergence and visualizations of MNIST autoencoder approximations.  For the convergence plots, the networks are trained for $50$ epochs and with recommended learning rate of $\gamma=10^{-3}$, batch size of $|\Tcal_k|=32$, regularization parameter $\alpha=10^{-10}$ for $\bftheta$, and 50,000 training images plus 10,000 for validation.  For ADAM, we train with three different regularization parameters for $\bfw$, $\lambda=10^{0}, 10^{-1}, 10^{-10}$.  When using {\tt slimTrain}, we automatically select the regularization parameters using sGCV with initial parameter $\Lambda_0=10^{-1}$ and choose a modest memory depth of $r=5$.  We display the DNN approximations after the first epoch below the convergence plots. The top row of MNIST images are, from left to right, $16$ test images, the approximation from the ADAM-trained networks with various regularization parameters on $\bfw$, and the approximation obtained from {\tt slimTrain}. The bottom row of images are the absolute differences (in log scale) between the network approximations and the true test images.  The value below the absolute difference images is the test loss over all 10,000 test images after the first epoch.}
\label{fig:autoencoderResultsFullData}
\end{figure}

The convergence results comparing {\tt slimTrain} and ADAM are presented in~\cref{fig:autoencoderResultsFullData}.  Here, we see that training with {\tt slimTrain} converges faster than ADAM in the first $10$ epochs and to a comparable lowest loss after $50$ epochs. Each training scheme forms an autoencoder that approximates the MNIST data accurately and generalizes well, even after the first epoch.  However, the absolute difference between the {\tt slimTrain} approximation and the true test images after the first epoch is noticeably less noisy than the ADAM-trained approximations after the first epoch, particularly for a poor choice of regularization parameter on $\bfw$ (e.g., $\lambda=10^0$).  We note that because we employ automatic regularization parameter selection, the performance of {\tt slimTrain} was nearly identical with different initial regularization parameters, $\Lambda_0$.  We display the case that produced slightly less oscillatory convergence.

\begin{figure}
\centering

\includegraphics[width=0.7\textwidth]{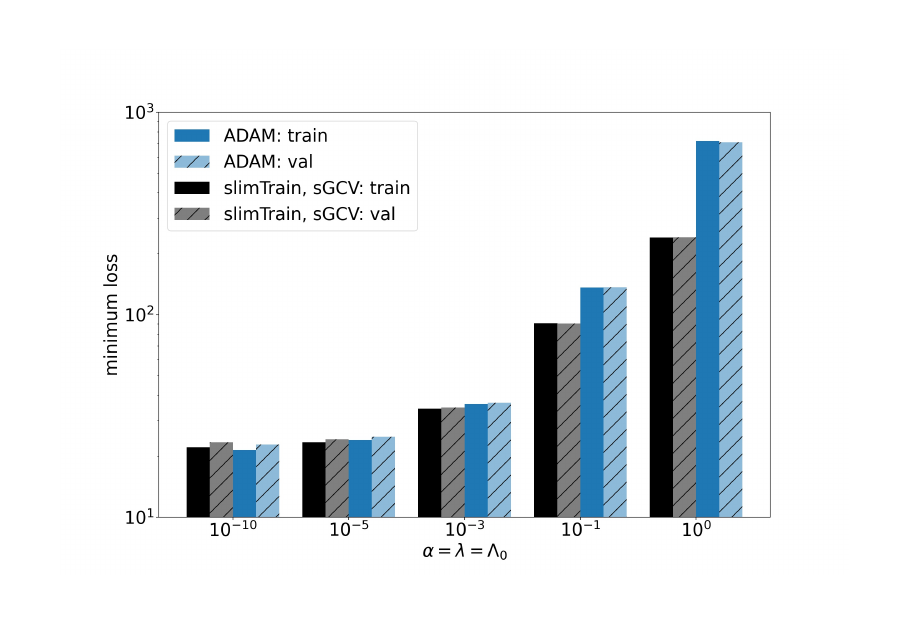}

\caption{Effect of regularization parameters on the minimum loss. We vary the regularization parameter $\alpha$ of $\bftheta$ for both ADAM (blue) and {\tt slimTrain} (black), the regularization parameter $\lambda$ on $\bfw$ for ADAM, and the initial regularization parameter $\Lambda_0$ on $\bfw$ for {\tt slimTrain}. For simplicity, we set the (initial) regularization parameters equal, $\alpha=\lambda=\Lambda_0$. The height of each bar is the training (solid) and validation (striped) loss for the network that obtained the lowest validation loss in $50$ epochs for the given hyperparameters.}
\label{fig:autoencoderAlpha}
\end{figure}

Using a good choice of the regularization parameter on the nonlinear weights ($\alpha=10^{-10}$) is partially responsible for the quality approximations obtained for each training method. The results in~\cref{fig:autoencoderAlpha} support our choice of a small regularization parameter on $\bftheta$. It can be seen that smaller regularization parameters on $\bftheta$ produce better DNN approximations.  When $\alpha$ is poorly-chosen (in this case, when $\alpha$ is large), {\tt slimTrain} produces a considerably smaller loss than training with ADAM. Hence, training with {\tt slimTrain} and sGCV can adjust to poor hyperparameter selection, even when those hyperparameters are not directly related to the regularization on $\bfw$. 

\begin{figure}
\centering

\includegraphics[width=0.9\textwidth]{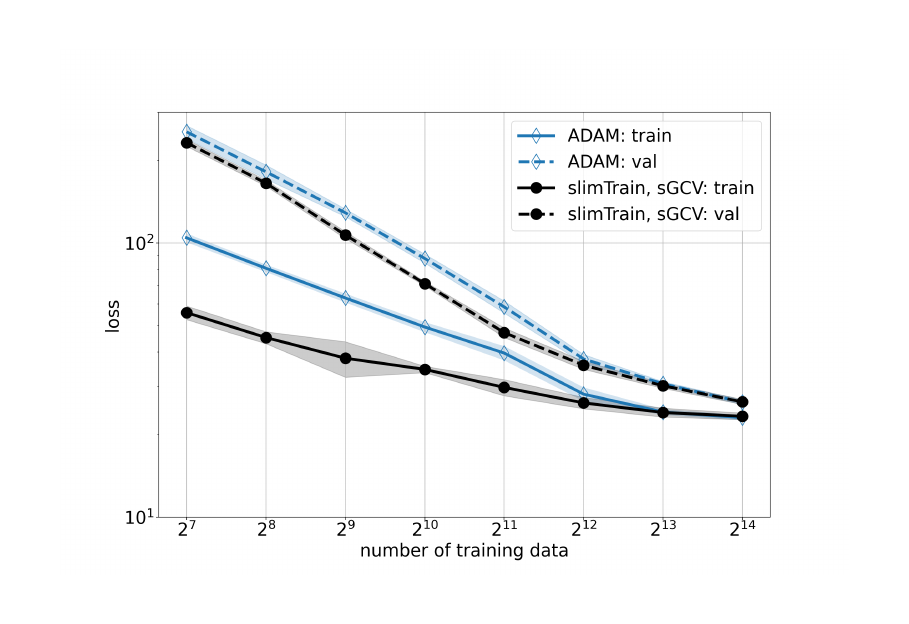}

\caption{Mean loss of MNIST autoencoder for small number of training data with a batch size of $32$.  All networks were trained for $50$ epochs for $10$ different weight initializations using the same hyperparameters as with the full training data in~\cref{fig:autoencoderResultsFullData}.  For each initialization, we choose the network that produced the minimal validation loss over $50$ epochs.  In the plot, each point denotes the mean loss over the $10$ runs and the bands depict one standard deviation from the mean.}	
\label{fig:autoencoderResultsSmallData}
\end{figure}

In addition to adjusting regularization parameters for the linear weights, we found that training with {\tt slimTrain} offers significant performance benefits in the limited-data setting; see~\cref{fig:autoencoderResultsSmallData}. When only a few training samples were used, training with {\tt slimTrain} produces a lower training and validation loss.  In the small training data regime, the optimization problem is more ill-posed and there are fewer network weight updates per epoch.  Hence, the automatic regularization selection and fast initial convergence of {\tt slimTrain} produces a more effective autoencoder.

Consistent with the results in our previous experiments, in the autoencoder example with a final convolutional layer, {\tt slimTrain} converges faster initially than ADAM to a good approximation and is less sensitive to the choice regularization on the nonlinear weights, $\bftheta$.   In the case of limited data, a common occurance for scientific applications, the training problem becomes more ill-posed.  Here, {\tt slimTrain} produces networks that fit and generalize better than ADAM.   By solving for good weights $\bfw$ and automatically choosing an appropriate regularization parameter at each iteration, {\tt slimTrain} achieves more consistent training performance for many different choices of hyperparameters.

% ====================================================================== %
% Section 5 - Conclusions
% ====================================================================== %
\section{Conclusions}
\label{sec:conclusions}

We address the challenges of training DNNs by exploiting the separability inherent in most commonly-used architectures whose output depends linearly on the weights of the final layer.  Our proposed algorithm, {\tt slimTrain}, leverages this separable structure for function approximation tasks where the optimal weights of the final layer can be obtained by solving a stochastic regularized linear least-squares problem. The main idea of {\tt slimTrain} is to iteratively estimate the weights of the final layer using the sampled limited-memory Tikhonov scheme {\tt slimTik}~\cite{chung2020sampled}, which is a state-of-the-art method to solve stochastic linear least-squares problems. By using {\tt slimTik} to update the linear weights, {\tt slimTrain} provides a reasonable approximation for the optimal linear weights and  simultaneously estimates an effective regularization parameter for the linear weights. The latter point is crucial -- {\tt slimTrain} does not require a difficult-to-tune learning rate and automatically adapts the regularization parameter for the linear weights, which can simplify the training process. In our numerical experiments, {\tt slimTrain} is less sensitive to the choice of hyperparameters, which can make it a good candidate to train DNNs for new datasets with limited experience and no clear hyperparameter selection guidelines. 

From a theoretical perspective, {\tt slimTrain} can be seen as an inexact version of the variable projection~\cite{GolubPereyra1973,OLearyRust2013} (VarPro) scheme extended to the stochastic approximation (SA) setting. Using this viewpoint, we show in ~\cref{sub:saslimtik} that we obtain unbiased gradient estimates for the nonlinear weights when the linear weights are estimated accurately. This motivates the design of {\tt slimTrain} as a tractable alternative to VarPro SA, which is infeasible as it requires re-evaluation of the nonlinear feature extractor over many samples after every training step. The computational costs of {\tt slimTrain} are limited as it re-uses features from the most recent batches and therefore adds little computational overhead; see~\cref{sec:implementation}. In addition, {\tt slimTrain} approximates the optimal linear weights obtained from VarPro, thereby reducing the bias introduced by the approximation when updating the nonlinear weights.

From a numerical perspective, the benefits of {\tt slimTrain}, and specifically automated hyperparameter selection, are demonstrated by the numerical experiments for both fully-connected and convolutional final layers.  In~\cref{sec:peaks}, we explore the relationship of the {\tt slimTrain} parameters, observing that memory depth and batch size play a crucial role in determining the ill-posedness of the least-squares problem to solve for the linear weights.  The regularization parameter adapts to the least-squares problem accordingly -- larger regularization parameters are selected when the problem is underdetermined. In~\cref{sec:surrogate}, we observe that {\tt slimTrain} is less sensitive to the choice of learning rate, outperforming the recommended settings for ADAM.  Again, the regularization parameters adapt to the learning rate -- larger parameters are chosen when the nonlinear weights change more rapidly.  In~\cref{sec:autoencoder}, we show that {\tt slimTrain} can be applied to a final convolutional layer and outperforms ADAM in the limited-data regime, which is typical in scientific applications.

% ====================================================================== %
% Acknowledgements and Bibliography
% ====================================================================== %
 
\section*{Acknowledgments}
This work was initiated as a part of the SAMSI Program on Numerical Analysis in Data Science in 2020. Any opinions, findings, and conclusions or recommendations expressed in this material are those of the authors and do not necessarily reflect the views of the National Science Foundation.

\bibliographystyle{siamplain}
\bibliography{references}

% ====================================================================== %
% Appendix
% ====================================================================== %
\appendix

% --------------------------------------------------------------------------------------------------------------------------- %
\section{Stochastic linear Tikhonov problem}
\label{sec:stochlinearTik}
In this section, we show that under certain assumptions, the stochastic Tikhonov-regularized least-squares problem \cref{eq:objFctnLinphi} has a closed form solution \cref{eq:What}.
Let us begin by defining $\bfmu_\bfy(\bftheta) = \bbE F(\bfy,\bftheta)$, $\bfSigma_\bfy(\bftheta) = \bbE (F(\bfy,\bftheta) -\bfmu_\bfy)(F(\bfy,\bftheta) -\bfmu_\bfy)\t$ and $\bfmu_\bfc(\bftheta) = \bbE \bfc$, $\bfSigma_\bfc = \bbE (\bfc-\bfmu_\bfc) (\bfc-\bfmu_\bfc)\t$. Then using the identity $$\bbE(\bfdelta\t \bfLambda \bfdelta) = \trace{\bfLambda\bfSigma_{\bfdelta}} + \bfmu_{\bfdelta}\t\bfLambda \bfmu_{\bfdelta},$$ where $\trace{\mdot}$ denotes the trace of a matrix, we have (sans constant from the regularization term for $\bftheta$)
\begin{align}
     \Phi(\bfW, \bftheta) &= \bbE \ \thf \norm[2]{\bfW F(\bfy,\bftheta) - \bfc}^2 + \tfrac{\lambda}{2} \norm[\rm F  ]{\bfW}^2\\
     &= \bbE \ \thf \left(\bfW F(\bfy,\bftheta) - \bfc\right)\t \left(\bfW F(\bfy,\bftheta) - \bfc\right)+ 
     \tfrac{\lambda}{2} \norm[\rm F  ]{\bfW}^2\\
     &= \bbE  \ \thf F(\bfy,\bftheta)\t\bfW\t\bfW F(\bfy,\bftheta) -  \bbE \bfc\t\bfW F(\bfy,\bftheta)  +  \thf \bbE \bfc\t\bfc + \tfrac{\lambda}{2} \norm[\rm F  ]{\bfW}^2\\
     &= \thf \trace{\bfW\t\bfW \bfSigma_{\bfy}(\bftheta)}  + \thf \bfmu_{\bfy}(\bftheta)\t \bfW\t\bfW \bfmu_{\bfy}(\bftheta) - \bbE \bfc\t\bfW F(\bfy,\bftheta)\\
     & \quad + \thf \trace{ \bfSigma_\bfc} +  \thf \bfmu_\bfc\t\bfmu_\bfc + \tfrac{\lambda}{2} \norm[\rm F  ]{\bfW}^2  
\end{align}
Notice that this function is quadratic in $\bfW$, and so for a given $\bftheta$ a minimizer \cref{eq:objFctnLinphi} can be found by differentiation.  That is,
\begin{equation}
    \rm D_\bfW  \Phi(\bfW,\bftheta) =  \bfW \bfSigma_{\bfy}(\bftheta) +  \bfW \bfmu_\bfy(\bftheta)\bfmu_\bfy(\bftheta)\t   + \lambda\bfW - \bbE \bfc F(\bfy,\bftheta)\t
\end{equation}
assuming we can switch order $\rm D \bbE = \bbE \rm D$. Now setting $\rm D_\bfW  \Phi = \bfzero$, we get
\begin{equation}
    \widehat\bfW(\bftheta) \left(\bfSigma_{\bfy}(\bftheta)  +  \bfmu_\bfy(\bftheta)\bfmu_\bfy(\bftheta)\t  + \lambda\bfI\right) = \bbE \bfc F(\bfy,\bftheta)\t
\end{equation}
and hence \cref{eq:What}.

% --------------------------------------------------------------------------------------------------------------------------- %
\section{Residual neural networks (ResNets)}
\label{app:resnet}
Residual neural networks (ResNets), among the most popular DNN architectures, are composed of layers of the form
	\begin{align}\label{eq:resnet}
	\bfu_0 &= \sigma(\bfK_{\rm in}\bfy + \bfb_{\rm in})\\
	\bfu_{j+1} &= \bfu_j + h\sigma(\bfK_j\bfu_j + \bfb_j) \text{ for }j=0,\dots, d-1.
	\end{align}
The architecture is defined by the width (the number of entries in the feature vectors $\bfu_j$), the depth (the number of layers $d$), and the step size $h > 0$.  The key property of ResNets is the identity mapping or skip connection which enables deeper, more expressive networks to be trained~\cite{he2016deep}. Recent work has interpreted ResNets as discretizations of continuous differential equations or dynamical systems~\cite{E2017} which have led to notions of stability~\cite{HaberRuthotto2017}, PDE-inspired architectures~\cite{RuthottoHaber2019}, and Neural ODEs~\cite{chen2018neural}.

In~\cref{sec:peaks}, we train a DNN to map $\bfy\in \Rbb^2$  to a scalar $c\in \Rbb$.  The feature extractor is a ResNet with a width of $w=8$ and a depth of $d=8$ corresponding to a final time of $T=5$ or equivalently with a step size of $h=5/8$. In~\cref{sec:surrogate}, we train a DNN to map $\bfy\in \Rbb^{55}$  to a scalar $c\in \Rbb^{72}$.  The feature extractor is a ResNet with a width of $w=16$ and a depth of $d=8$ corresponding to a final time of $T=5$ or equivalently with a step size of $h=5/8$.  In both experiments, we use the smooth hyperbolic tangent activation function, $\sigma(x) = \tanh(x)$.

% --------------------------------------------------------------------------------------------------------------------------- %
\section{Autoencoder architecture}
\label{app:autoencoder}
We adapt the MNIST autoencoder from~\cite{Malmaud2018}.  The autoencoder consists of two convolutional neural networks with a user-defined width $w$ and intrinsic dimension $d$.  The width controls the number of convolutional filters used and the intrinsic dimension is the size of the low-dimensional embedding. The architecture is described in~\cref{tab:autoencoderArchitecture}.

\setlength\extrarowheight{5pt}

\begin{table}
    \centering
    \caption{Autoencoder architecture with a width $w = 16$ and intrinsic dimension $d$. For the convolutional layers, $s$ is the stride and $p$ is the padding.  The layer ConvT indicates a transpose convolution. The dashed line indicates the separable final layer.}
    \label{tab:autoencoderArchitecture}
    {\tiny
    \addtolength{\tabcolsep}{-1pt}    
    \begin{tabular}{|l|c|c|c|c|c|}
    \hline
    & Layer Type & Description & \# Feat. Out & \# Weights & $w=16$, $d=50$ \\[1ex]
    \hline\hline
    
    \multirow{3}{*}{\rotatebox{90}{Enc}} 
    
        & Conv. + ReLU & $w$, $4\times 4\times 1$ filters, $s=2$,  $p=1$
                             & $14 \times 14 \times w$ & $16w + w$ & 272\\[1ex]
        & Conv. + ReLU & $2w$, $4\times 4\times w$ filters, $s=2$,  $p=1$
                             & $7 \times 7 \times 2w$ & $32w^2 + 2w$ & $8,224$\\[1ex]
        & Affine & $d\times (49 \cdot 2w)$ matrix + $d\times 1$ bias
                             & $d\times 1$ & $98wd + d$ & $78,450$ \\[1ex]
    \hline
    \multirow{4}{*}{\rotatebox{90}{Dec}} 
        & Affine & $(49\cdot 2w)\times d$ matrix + $(49\cdot 2w)\times 1$ bias
                             & $98w\times 1$ & $98wd+ 98w$ & $79,968$ \\[1ex]
        & Batch Norm & \----
                             & \---- & \---- & \---- \\[1ex]
        & ConvT. + ReLU & $w$, $4\times 4\times 2w$ filters, $s=2$,  $p=1$ 
                             & $14 \times 14 \times w$ & $32w^2 + w$ & 8,208\\[1ex]
            \cdashline{2-6}         
        & ConvT. + ReLU & $1$, $4\times 4\times w$ filter, $s=2$,  $p=1$
                             & $28 \times 28 \times 1$ & $16w + 1$ & 257\\[1ex]
        \hline \hline
    \multicolumn{4}{|r|}{Total} & \---- &$175,122+ 257$\\[1ex]
    \hline

    \end{tabular}
    }

\end{table}

The final layer is a (transposed) convolution, denoted in \cref{sec:implementation} as $\bfK(\cdot): \Rbb^{|\bfw|} \to \Rbb^{\nFeatIn\times \nFeatOut}$.  Note that $\nFeatIn > \nFeatOut$ in our case. As we did in~\cref{sub:slimTik}, we can express the operation of $\bfK(\bfw)\in \Rbb^{\nFeatIn\times \nFeatOut}$ on the output features $\bfZ_k(\bftheta)\in \Rbb^{\nFeatOut \times |\Tcal_k|}$ as a linear operator applied to the weights $\bfw$; that is,
    \begin{align*}
        \bfK(\bfw)\bfZ_k(\bftheta)
        \qquad \begin{array}{c}
      \xrightarrow[\hspace{1cm}]{\text{de-conv}}\\[-1em] 
      \xleftarrow[\text{conv}]{\hspace{1cm}} \end{array} \qquad
      \bfA_k(\bftheta) \bfw.
    \end{align*}
The matrix $\bfA_k(\bftheta) \in \Rbb^{|\Tcal_k|\nFeatIn \times |\bfw|}$ has known structure.  In particular, each column of $\bfA(\bftheta)$ contains a shifted copy of $\text{vec}(\bfZ_k(\bftheta))$.  Na\"{i}vely, we can form each column of $\bfA_k(\bftheta)$ explicitly by applying the (transposed) convolution operator to ``standard basis'' filters. Specifically, the $j$-th column of $\bfA_k(\bftheta)$ is $\text{vec}(\bfK(\bfe_j)\bfZ_k(\bftheta))$ where $\bfe_j\in \Rbb^{|\bfw|}$ is the $j$-th unit vector. In our implementation, we construct $\bfA_k(\bftheta)$ by recognizing that the samples and the channels of $\bfZ_k(\bftheta)$ are independent, requiring fewer evaluations of the (transposed) convolution operator. Our code is available at \url{https://github.com/elizabethnewman/slimTrain}.

\end{document}